\documentclass[letterpaper]{article} 
\usepackage{aaai2026}  
\usepackage{times}  
\usepackage{helvet}  
\usepackage{courier}  
\usepackage[hyphens]{url}  
\usepackage{graphicx} 
\usepackage{subcaption}
\usepackage{natbib}  
\usepackage{caption} 
\usepackage{amsmath}
\usepackage{multirow}
\usepackage{algorithm}
\usepackage{algorithmic}

\usepackage{newfloat}
\usepackage{listings}
\DeclareCaptionStyle{ruled}{labelfont=normalfont,labelsep=colon,strut=off} 
\floatstyle{ruled}
\newfloat{listing}{tb}{lst}{}
\floatname{listing}{Listing}
\title{Towards Better IncomLDL: We Are Unaware of Hidden Labels in Advance}
\author{
    Jiecheng Jiang\textsuperscript{\rm 1,2}\equalcontrib,
    Jiawei Tang\textsuperscript{\rm 1}\equalcontrib,
    Jiahao Jiang\textsuperscript{\rm 1},
    Hui Liu\textsuperscript{\rm 3},
    Junhui Hou\textsuperscript{\rm 4}\thanks{Corresponding authors.},
    Yuheng Jia\textsuperscript{\rm 1,3,5}\footnotemark[2]
}
\affiliations{
    \textsuperscript{\rm 1}School of Computer Science and Engineering, Southeast University, Nanjing 210096, China\\
    \textsuperscript{\rm 2}School of Computing, National University of Singapore, Singapore\\
    \textsuperscript{\rm 3}School of Computing Information Sciences, Saint Francis University, Hong Kong, China\\
    \textsuperscript{\rm 4}Department of Computer Science, City University of Hong Kong, Hong Kong, China\\
    \textsuperscript{\rm 5}Key Laboratory of New Generation Artificial Intelligence Technology and Its \\Interdisciplinary Applications (Southeast University), Ministry of Education, China

    \{jcjiang, jwtang, jhjiang, yhjia\}@seu.edu.cn, hliu99-c@my.cityu.edu.hk, jh.hou@cityu.edu.hk
}

\usepackage{bibentry}

\begin{document}

\maketitle

\begin{abstract}
Label distribution learning (LDL) is a novel paradigm that describe the samples by label distribution of a sample. However, acquiring LDL dataset is costly and time-consuming, which leads to the birth of incomplete label distribution learning (IncomLDL). All the previous IncomLDL methods  set the description degrees of ``missing" labels in an instance to 0, but remains those of other labels unchanged. This setting is unrealistic because when certain labels are missing, the degrees of the remaining labels will increase accordingly. We fix this unrealistic setting in IncomLDL and raise a new problem: LDL with hidden labels (HidLDL), which aims to recover a complete label distribution from a real-world incomplete label distribution where certain labels in an instance are omitted during annotation. To solve this challenging problem, we discover the significance of proportional information of the observed labels and capture it by an innovative constraint to utilize it during the optimization process. We simultaneously use local feature similarity and the global low-rank structure to reveal the mysterious veil of hidden labels. Moreover, we \textbf{theoretically} give the recovery bound of our method, proving the feasibility of our method in learning from hidden labels. Extensive recovery and predictive experiments on various datasets prove the superiority of our method to state-of-the-art LDL and IncomLDL methods.
\end{abstract}


\begin{links}
    \link{Code}{https://github.com/Trisitana/HidLDL}
\end{links}

\section{Introduction}

Learning with ambiguity has gained notable attention in machine learning community recently \cite{LDL}. In classical machine learning, single-label learning refers to the process of assigning exactly one label to an instance. Multi-label learning \cite{MLL,MLL2,liu2021emerging}, on the other hand, assigns multiple labels to a single instance. However, in practical applications, for the same instance, we need to consider the relative importance of their labels. Label distribution learning (LDL) \cite{LDL} is a novel machine learning paradigm to solve the above issue by modeling the importances of multiple labels for each instance. In recent years, LDL have been widely used in many applications, such as age estimation \cite{app1,app1_2}, emotion recognition \cite{app2,app2_2}, facial beauty \cite{app3}, crowd opinion prediction \cite{app4}, medical diagnosis \cite{app5},  and scene text detection \cite{app6_2,app6} etc.

\begin{figure*}[t]
    \centering
    \begin{subfigure}[b]{0.22\textwidth}
        \centering   \includegraphics[width=\textwidth]{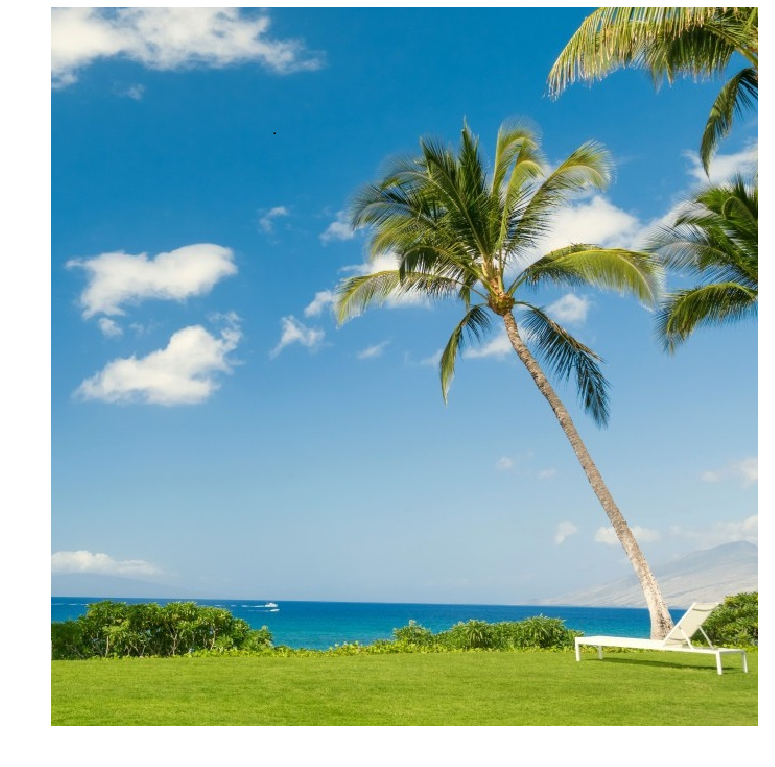}
        \caption{image}
        \label{instance_example}
    \end{subfigure}
    \hfill
    \begin{subfigure}[b]{0.22\textwidth}
        \centering      
        \includegraphics[width=\textwidth]{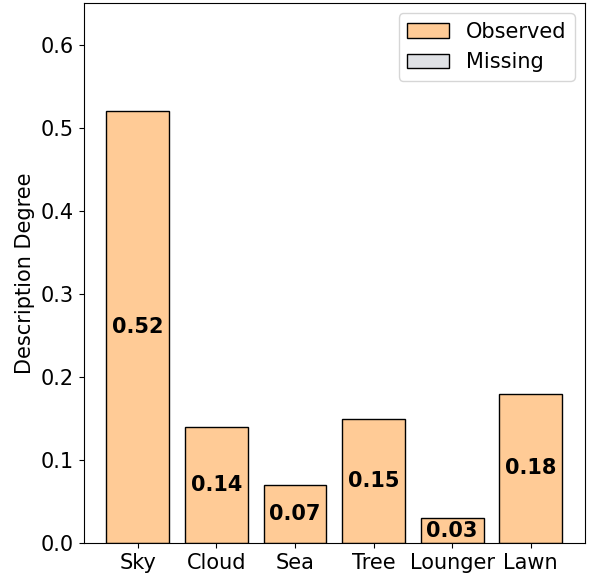}
        \caption{LDL}
        \label{LDL_example}
    \end{subfigure}
    \hfill
    \begin{subfigure}[b]{0.22\textwidth}
        \centering     \includegraphics[width=\textwidth]{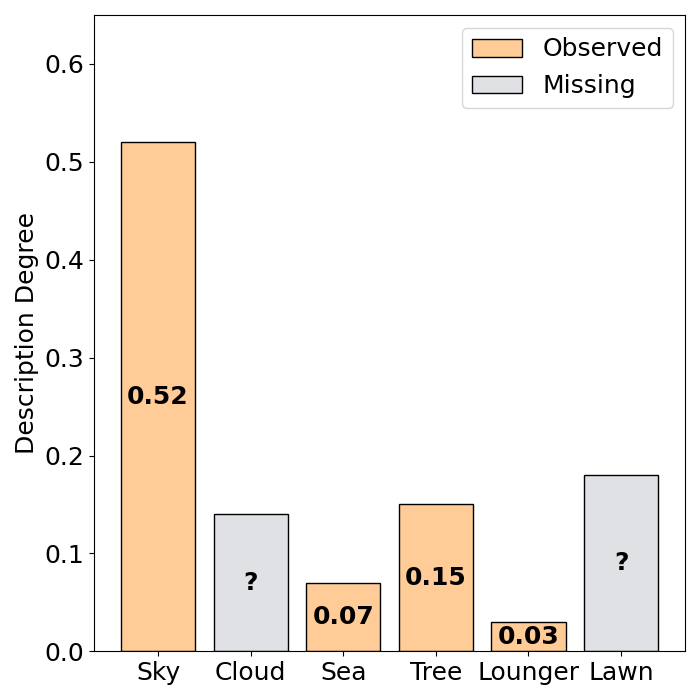}
        \caption{IncomLDL}
        \label{IncomLDL_example}
    \end{subfigure}
    \hfill
    \begin{subfigure}[b]{0.22\textwidth}
        \centering        \includegraphics[width=\textwidth]{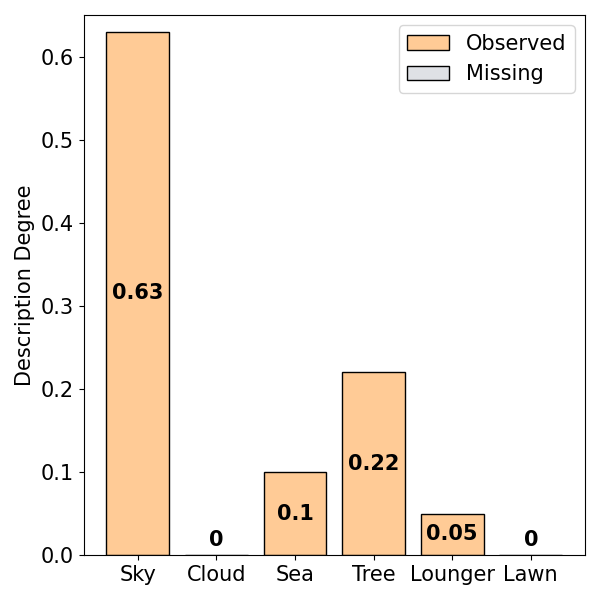}
        \caption{HidLDL}
        \label{HidLDL_example}
    \end{subfigure}
    \caption{(a) A scene image containing 6 elements. (b) The label distribution (LD) of the image. (c) The observed LD in IncomLDL, with gray indicating missing (unobserved) labels. The sum of the description degrees of the observed labels is not 1. (d) The observed LD in HidLDL. For the unobserved labels in (c), the corresponding description degrees are 0, and the sum of the description degrees of the observed labels is 1. HidLDL is more intuitive and realistic, as the observed labels should occupy all description degree.}
    \label{fig:example}
    \vspace{-4mm}
\end{figure*}


In LDL, each instance $\boldsymbol{x}$ is annotated by a real value vector $d_{\boldsymbol{x}}$, i.e., label distribution (LD). Each component of the LD vector $d_{\boldsymbol{x}}^y$ represents the description degree of the label $y$ to the instance $\boldsymbol{x}$. Assume $d_{\boldsymbol{x}}^y \in[0,1]$ and the label set is complete, which means that all the labels in the set can always fully describe the instance. Then, we have $\sum_y d_{\boldsymbol{x}}^y=1$.

Despite LDL being successfully applied to many scenarios in recent years, existing LDL methods are incapable of handling incomplete supervised information. In practice, the description degrees are often provided by human annotators, which can be extremely costly in terms of labor and time when dealing with a large dataset. Meanwhile, data corruption and incomplete annotations may also occur. To this end, incomplete label distribution learning (IncomLDL) \cite{inLDL} was proposed to address the problem of incomplete supervised data.  In the previous research of IncomLDL, a predefined set of labels is given. The description degrees of some labels for a given instance are randomly set to 0, and those
of other labels still remain unchanged.

However, the setting of IncomLDL is \textbf{unreasonable} and \textbf{unrealistic}. In real situations, when
certain labels are missing (their description degrees set
to 0 in IncomLDL), degrees of remaining labels should increase accordingly instead of remaining unchanged. Below, we use an illustration to better explain this unreasonable setting inherited in IncomLDL.
 
Fig.~\ref{instance_example} shows a scene image containing 6 elements, and its LD $d_{\boldsymbol{x}} = \{0.52, 0.14, 0.07, 0.15, 0.03, 0.18\}$ is presented in Fig.~\ref{LDL_example}. In IncomLDL, some of the description degrees are randomly set to 0 so that incomplete data of LD are formed into $d_{\boldsymbol{x}} = \{0.52, 0, 0.07, 0.15, 0.03, 0 \}$ (Fig.~\ref{IncomLDL_example}). This is an approximation of real-world missing data. However, unless a portion of the data in the database is lost, this setting has a significant distance from the real world. Suppose we are the annotators. When annotating the image in Fig.~\ref{instance_example}, we accidentally omit the labels for ``cloud'' and ``lawn'' (such omitted labels are referred to hereafter as hidden labels). In this case, the remaining labels ought to form a new comprehensive description of the instance, and the sum of their description degrees should equal 1. As illustrated in Fig.~\ref{HidLDL_example}, with ``cloud'' and ``lawn'' hidden from the annotator’s view, the description degrees of the other labels will increase to compensate and still manage to provide a complete description. Therefore, in IncomLDL, it is not a viable practice to simply set description degrees of hidden labels to 0 while keeping the description of the observed labels unchanged.


Based on the aforementioned illustration, we contend that current research on IncomLDL does not adequately reflect the true nature of incompleteness and leaves considerable room for refinement. We propose a new setting called LDL with hidden labels (HidLDL), which is closer to real situations. \textbf{We then propose a novel method that pays close attention to the crucial proportion information of observed labels, which is our most outstanding innovation.} By exploiting local feature similarity and the global low-rank structure to capture both local and global dependencies, we strive for further improvement. Moreover, we \textbf{theoretically}
give the recovery bound of our method, proving the feasibility of our method in learning from hidden labels. Experimental results have shown that our method has significantly better performance compared to existing methods. The main contributions can be summarized as follows. 

\begin{itemize}
    \item We notice the unreasonable aspect of all previous incomplete LDL research,
    and propose a novel setting, abbreviated as HidLDL, which is closer to real situations.
    \item We discover the need to fully utilize the proportion information of observed labels in the new HidLDL problem. During optimization, apart from simultaneously using both local feature similarity and global low-rank structure, we innovatively use constraints to ensure the correct use of proportion information. 
    
    \item We theoretically give the recovery bound of our method, in which the optimization objective and constraints of our method help determine the recovery bound, proving the feasibility of our method in learning from hidden labels.
   \item We empirically verify the
   effectiveness of our method on 12 real-world datasets in both recovery and predictive tasks.
    Experimental results show that our method outperforms state-of-the-art LDL and IncomLDL methods, with the improvements being statistically significant in numerous cases.
    
\end{itemize}


\section{Related Work}

\subsection{Label Distribution Learning}

Label distribution learning (LDL) \cite{rankmatch, concentration, DLDL, instancedepend, biased} was first proposed to solve the facial age estimation problem. Later on,  \cite{LDL} discovered that in some real applications, the distribution across all labels is more desirable than the association of a single label to an instance. Since LDL was introduced, numerous methods have been proposed. These methods can be roughly divided into three categories: problem transformation, algorithm adaptation and specialized algorithms. 
Yang \cite{yang} proposed a deep learning based LDL method. Two classic methods SA-IIS and BFGS-LLD were then proposed in \cite{LDL}. LDLLC \cite{LC} maps label distances to parameter matrix column distances and uses Pearson correlation to capture global label relationships. GD-LDL-SCL and Adam-LDL-SCL \cite{LS} were designed to exploit local correlation in different groups which helps solve LDL problems. LDL-LDM \cite{LDL-LDM} was proposed to exploit label correlation maniford which is helpful to alleviate the overwhelm of LDL outspace. In \cite{ordinal}, the authors proposed an auxiliary MLL process integrated into the LDL framework, which captures low-rank label correlations within this additional MLL component. Also, a new paradigm based on LDL, called OLDL \cite{ordinal}, was proposed focusing on  the internal sequential patterns of labels.

\subsection{Incomplete  Label Distribution Learning}
The above mentioned LDL methods all assume that in the training set, each sample is annotated with a full label distribution. But in reality, the labels of some samples are either not observed, or it is expensive to annotate  complete training samples \cite{complementary, ensemble, semisupervise}.
To this end, incomplete label distribution learning (IncomLDL) was introduced by \cite{inLDL} to address annotation incompleteness. To solve the LDL problem when given incomplete supervised information, they suggested a low-rank model to utilize the correlation between labels and proposed IncomLDL-prox and IncomLDL-admm. Adapting existing LDL methods to the scenario of missing annotations is another intuitive strategy for addressing incomplete LDL tasks. In  \cite{WInLDL}, the authors designed a weighting scheme of label degrees to utilize prior information of label distribution itself. In  \cite{LDL-LDM}, the authors contended that the assumption of low-rank might not always be valid. They proposed an alternative model where the predictions are constrained to lie on a shared manifold, with the manifold’s structure capturing the correlations between labels. Recently, in \cite{inLDL-LCD}, the authors proposed an Incomplete Label Distribution Learning method via correlation decomposition by assuming a globally low-rank structure and a locally sparse structure. They developed an alternating solution with the accelerated proximal gradient descent method for optimization (IncomLDL-LCD). 


\section{LDL with Hidden Labels}

\subsection{Problem Setting}
Let $\boldsymbol{x}\in R^d$ denote a feature vector, $\mathbf{X}=\left[\boldsymbol{x}_1;\boldsymbol{x}_2;\ldots;\boldsymbol{x}_n\right]\in R^{n\times d}$ denote the feature matrix and $\mathcal{Y}=\left\{y_{1}, y_{2}, \ldots, y_{m}\right\}$ denote the complete set of labels, where $d$, $n$ and $m$ represent the dimension of features, the number of samples, and the number of labels. In LDL, the description degree of the label $y$ to the instance $\boldsymbol{x}$ is denoted by $d_{\boldsymbol{x}}^{y}$, which satisfies $d^{y}_{\boldsymbol{x}} \in \left[0,1\right]$ and $\sum_y d^{y}_{\boldsymbol{x}}=1$, the label distribution of $\boldsymbol{x}_i$ is denoted by $\boldsymbol{d}^{g}_i=\left(d^{y_{1}}_{\boldsymbol{x}_i},d^{y_{2}}_{\boldsymbol{x}_i},\ldots,d^{y_{m}}_{\boldsymbol{x}_i}\right) \in R^m$, and the ground-truth label distribution matrix is denoted by $\mathbf{D}^{g}=\left[\boldsymbol{d}^{g}_1;\boldsymbol{d}^{g}_2;\ldots;\boldsymbol{d}^{g}_n\right]\in R^{n\times m}$.

In IncomLDL \cite{inLDL}, for each sample $\boldsymbol{x}$, there exists a possibility that some entries in $d^{y}_{\boldsymbol{x}}$ may not be observed, aiming to simulate the real-world scenarios. Specially, let $\mathbf{M} \in R^{n \times m}$ denote the indices of the observed labels in $\mathbf{D}^{g}$:
\begin{equation}
\mathit{M}_{i j}= \begin{cases}1 & \text { if $D^{g}_{ij}$ is observed,} \\ 0 & \text { otherwise.}\end{cases}
\end{equation}
Then, the observed label distribution matrix in IncomLDL is defined as, 
\begin{equation}
\mathit{D^{o}_{\mathit{ij}}} = 
\begin{cases}
\mathit{D^{\mathit{g}}_{\mathit{ij}}} & \text{if } \mathit{M_{\mathit{ij}}} = 1, \\
0 & \text{if } \mathit{M_{\mathit{ij}}} = 0.
\end{cases}
\label{normalize}
\end{equation}
However, this approach is not reasonable and has been discussed in detail in the previous context.


Therefore, we propose a new setting called HidLDL. Specifically, in HidLDL, due to some labels not being observed at the beginning, in the eyes of the annotator, other labels will occupy the degree of these hidden labels. We assume that the degrees of the observed labels in $\mathbf{D}^{g}$ will amplify proportionally. Based on this instinctive assumption, a more reasonable $\mathbf{D}^{o}$ is defined as, 
\begin{equation}
\mathit{D^{o}_{\mathit{ij}}} = 
\begin{cases}
\displaystyle \frac{\mathit{D^{\mathit{g}}_{\mathit{ij}}}}{\sum\limits_{k=1}^{m} \mathit{D^{\mathit{g}}_{\mathit{ik}}} \cdot \mathit{M_{\mathit{ik}}}} & \text{if } \mathit{M_{\mathit{ij}}} = 1, \\
\quad \quad 0 & \text{if } \mathit{M_{\mathit{ij}}} = 0.
\end{cases}
\label{normalize}
\end{equation}
The goal of HidLDL is to recover a label distribution matrix $\mathbf{D}=\left[\boldsymbol{d}_1;\boldsymbol{d}_2;\ldots;\boldsymbol{d}_n\right]\in R^{n\times m}$, which is as close as possible to the ground-truth label distribution. Due to the inaccuracy of the observed label distribution, HidLDL receives more ambiguity  compared to IncomLDL and requires the use of new approach to solve it.

\subsection{Proposed Method}
In order to recover hidden labels, a natural idea is to learn from other instances containing these hidden labels. Based on the smoothness assumption \cite{smooth}, in the feature space, instances close to each other tend to share a common label. Even if the observed degree of labels is not accurate in HidLDL, the nearby points can still learn from each other. 

\subsubsection{Graph Dependency} To capture graph dependency, we use K-nearest neighbors algorithm to determine whether $\boldsymbol{x}_i$ and $\boldsymbol{x}_j$ are connected. Setting the neighbor number to the label number $m$, $\boldsymbol{x}_i$ and $\boldsymbol{x}_j$ are connected if $\boldsymbol{x}_i$ is among K-nearest neighbors of $\boldsymbol{x}_j$ or vice versa. At first, the similarity matrix $\mathbf{A}$ is set to $\mathbf{0}_{n \times n}$, indicating that there are no initial dependencies. Then, if $x_i$ and $x_j$ are connected, $A_{ij}$ is updated in Eq. (\ref{Gaussian kernel}): 
\begin{equation}
\label{Gaussian kernel}
A_{i j}=\exp \left(-\frac{\left\|\boldsymbol{x}_i-\boldsymbol{x}_j\right\|^2_{2}}{2 \sigma^2}\right),
\end{equation} where ${\sigma}$ is the bandwidth parameter of the Gaussian kernel. If $\boldsymbol{x}_i$ and $\boldsymbol{x}_j$ are closely connected, then $\boldsymbol{d}_i$ and $\boldsymbol{d}_j$ should be close to each other. This idea leads to the following optimization objectives $\mathit{\Omega}(\mathbf{D})$: 

\begin{equation}
\begin{aligned}
\mathit{\Omega}(\mathbf{D}) =\sum_{i, j} A_{i j}\left\|\boldsymbol{d}_i-\boldsymbol{d}_j\right\|^2_{2} = \operatorname{tr}\left(\mathbf{D}^{\top} \mathbf{G} \mathbf{D}\right),
\end{aligned}
\end{equation}
where $\mathbf{G} = \hat{\mathbf{A}}-\mathbf{A}$ is the graph Laplacian and $\hat{\mathbf{A}}$ is a diagonal matrix whose elements are $\hat{A}_{i i}=\sum_{j=1}^n A_{i j}$.

\subsubsection{Low-rank Assumption} Global low-rank assumption has good performance in handling incomplete supervision information \cite{inLDL}. To further mine the correlation between labels, we assume that $\mathbf{D}$ is low-rank, i.e., we hope that the trace norm of the recovered matrix $\mathbf{D}$ is as small as possible. Therefore, the trace norm of $\mathbf{D}$ (i.e., $\|\mathbf{D}\|_*$) is added as an optimization term, which is the sum of all singular values of the matrix.

\subsubsection{Proposed Proportional Constraint} The most important thing in HidLDL is how to utilize the observed matrix $\mathbf{D}^{o}$. In  \cite{WInLDL}, Frobenius norm is used to minimize the difference in those observed positions, however, in HidLDL, this may not work because observed degrees of labels no longer directly reflect their final description of the instance. Specifically, we must only use the proportional relationship between the degrees of the observed labels to avoid introducing more noise. Even if we try to incorporate this proportion information of $\mathbf{D}$ into the final optimization objective, it not only introduces additional hyperparameters but also brings in more non-smooth terms. Thus, we define the following constraint set:

\begin{equation}
\begin{aligned}
Cons = \left\{\mathbf{Z} \in R^{n \times m} \mid \forall i \in \{1, 2, \ldots, n\}, \exists k_i \in R 
\right.\\
\qquad \text{s.t.} \ k_i \cdot \mathbf{d}^{o}_{i}=\mathbf{z}_{i} \odot \mathbf{m}_{i}\left.\right\},
\end{aligned}
\end{equation}
where the \textit{i}-th row of $\mathbf{Z}$ is denoted as $\mathbf{z}_{i}$, so is $\mathbf{m}_{i}$ and $\mathbf{d}^{o}_{i}$. 
We constrain $\mathbf{D} \in Cons$. We expect that each row of the recovered label matrix, after masking, is proportional to the initial observed matrix, with a scaling coefficient of $k_{i}$. This setting makes sense even when $\mathbf{D}^{o}_{ij}$ is $0$, thus increasing the generality of our model.

Note that the recovered label distribution needs to be in the probability simplex. We add constraints $\mathbf{D}\times \mathbf{1}_{m} = \mathbf{1}_{n}$ and $\mathbf{D}\geq  \mathbf{0}_{n\times m}$ to guarantee that all entries are negative and the sum of each row's entries is 1. Considering all the factors mentioned above, the final objective function of our method can be written as follows:
\begin{equation}
\label{final_object}
\begin{aligned}
& \min _{\mathbf{D}} \quad\frac{1}{2} \operatorname{tr}\left(\mathbf{D}^{\top} \mathbf{G}\mathbf{D}\right)+\alpha\|\mathbf{D}\|_* \\
& \text { s.t. } \quad\mathbf{D}\times \mathbf{1}_{m} = \mathbf{1}_{n} ,\ \mathbf{D}\geq\mathbf{0}_{n \times m}, \ \mathbf{D} \in Cons,
\end{aligned}
\end{equation}
where $\operatorname{tr}(\cdot)$ denotes the trace of a matrix, $\|\cdot\|_*$ denotes the trace norm of the matrix, and $\alpha$ serves as a regularization parameter that controls the trade-off between graph dependency and the trace norm.

\subsection{Theoretical Analysis}

In this section, we provide the recovery bound of our method.

\textbf{Lemma 1.} \textit{For $i$-th instance, the recovery bound of it can be represented by the error between the ground-truth scaling coefficient $k^g_i=\frac{d^o_{ij}}{d^g_{ij}}$ and the recovered scaling coefficient $k_i$. So, the target is to prove that}

\begin{equation}
\small
\label{Lemma1}
    (k^g_i-k_i)^2\leq \epsilon_i,
\end{equation}
\textit{where $\epsilon_i$ is a small number.}

\textbf{Theorem 1.} \textit{Under certain assumptions, the error of the scaling coefficients can be expressed as}

\begin{equation}
\small
\label{Theorem1}
    k^g_i-k_i \leq \frac{1-\sum_{k=1}^m D_{ik}M_{ik}}{\sum_{k=1}^m D^g_{ik}M_{ik}}.
\end{equation}

The detailed proof of \textbf{Theorem 1} is given in section \textbf{F} of the appendix. Let $\sigma_i=1-\Sigma_k D_{ik}M_{ik}$. In view of the definition of $M_{ik}$, $\sigma_i$ represents \textbf{the sum of the labels in $i$-th instance that are masked}, and the recovery bound of $i$-th instance $\epsilon_i$ becomes

\begin{equation}
\small
\label{Conclusion}
    \epsilon_i\leq\frac{\sigma^2}{(\Sigma_k D^g_{ik}M_{ik})^2}.
\end{equation}

From Eq. \eqref{Conclusion}, we can observe that its denominator is strictly greater than zero but less than one. And the smaller the $\sigma_i$ is, the tighter the recovery bound $\epsilon_i$ is. When $\sigma_i$ tends to zero, $\epsilon_i$ also tends to zero. This degrades the hidden label LDL problem to the learnable traditional LDL problem, which is an intuitive conclusion. \textit{Combining all the proofs together, we conclude that LDL with hidden labels through our method is theoretically feasible.}

\subsection{Optimizing using ADMM}
Considering that the optimization objective contains complex constraints and non-smooth terms, we use ADMM (Alternating Direction Method of Multipliers)  \cite{ADMM} to solve Eq. (\ref{final_object}). By breaking Eq. (\ref{final_object}) into smaller pieces, each of which will be easier to handle. Specifically, Eq. (\ref{final_object}) can be converted into the following equivalent forms by replacing the matrix $\mathbf{D}$ with $\mathbf{A}$ and $\mathbf{B}$:
\begin{equation}
\label{equivalent_object}
\begin{aligned}
&\min _{\mathbf{A,B,D}} \quad\frac{1}{2} \operatorname{tr}\left(\mathbf{D}^{\top} \mathbf{G}\mathbf{D}\right)+\alpha\|\mathbf{A}\|_* \\
&\text { s.t. } \quad\mathbf{D}\times \mathbf{1}_{m} = \mathbf{1}_{n} , \ \mathbf{D}\geq\mathbf{0}_{n \times m}, \ \mathbf{B} \in Cons,\\
& \quad \hspace{1.8em} \mathbf{D} - \mathbf{A} = \mathbf{0},\ \mathbf{D}-\mathbf{B}=\mathbf{0}.
\end{aligned}
\end{equation}
Then, the solution of Eq. (\ref{equivalent_object}) can be accomplished by solving the following augmented Lagrange multiplier equation:
\begin{equation}
\label{lagrange_multiplier_equation}
\begin{aligned}
& \min _{\mathbf{A}, \mathbf{B}, \mathbf{D}} \ \frac{1}{2}\operatorname{tr}\left(\mathbf{D}^{\top} \mathbf{G} \mathbf{D}\right)+\alpha\|\mathbf{A}\|_* \\
& \qquad \quad+  \left\langle\mathbf{\Lambda}, \mathbf{D}-\mathbf{A}\right\rangle+\frac{\rho}{2}\|\mathbf{D}-\mathbf{A}\|_F^2 \\
& \qquad \quad+  \left\langle\mathbf{\Lambda'}, \mathbf{D}-\mathbf{B}\right\rangle+\frac{\rho}{2}\|\mathbf{D}-\mathbf{B}\|_F^2\\
&\text { s.t. } \quad \mathbf{D} \times \mathbf{1}_{m} = \mathbf{1}_{n} , \ \mathbf{D}\geq\mathbf{0}_{n \times m}, \ \mathbf{B} \in Cons,\\
\end{aligned}
\end{equation}
where $\mathbf{\Lambda} \in R^{n \times m}$ and $\mathbf{\Lambda'} \in R^{n \times m}$ are the Lagrange multipliers, $\rho \geq 0$ is a penalty parameter, and $\langle\cdot, \cdot\rangle$ returns the inner product of two matrices. Although $\rho$ can increment in each loop, in our paper, we fix $\rho$ at $2$, which is sufficient for the objective to converge. To solve Eq. (\ref{lagrange_multiplier_equation}), the ADMM iteratively solves the following subproblems.

1) $\mathbf{D}$ Subproblem: Removing irrelated terms regarding $\mathbf{D}$, $\mathbf{D}$ subproblem is written as:
\begin{equation}
\label{op-D}
\begin{aligned}
&\min_{\mathbf{D}} \ \frac{1}{2} \operatorname{tr}\left(\mathbf{D}^{\top} \mathbf{G} \mathbf{D}\right) + \left\langle \mathbf{\Lambda} \cdot \mathbf{D} - \mathbf{A} \right\rangle + \frac{\rho}{2} \|\mathbf{D} - \mathbf{A}\|_F^2 \\
&\hspace{1.8em} + \left\langle \mathbf{\Lambda'} \cdot \mathbf{D} - \mathbf{B} \right\rangle + \frac{\rho}{2} \|\mathbf{D} - \mathbf{B}\|_F^2\\
&\text { s.t. } \quad\mathbf{D}\times \mathbf{1}_{m} = \mathbf{1}_{n} , \ \mathbf{D}\geq\mathbf{0}_{n \times m}.
\end{aligned}
\end{equation}
We use projected gradient descent (PGD) to solve
Eq. (\ref{op-D}). The gradient of the objective with respect to $\mathbf{D}$ is:
\begin{equation}
\label{SGD}
\nabla_\mathbf{D}=\mathbf{G}\mathbf{D}+\mathbf{\Lambda}+\mathbf{\Lambda'}+\rho(\mathbf{D} - \mathbf{A})+\rho(\mathbf{D} - \mathbf{B}).
\end{equation}
We use Stochastic Gradient Descent (SGD) to update $\mathbf{D}$ one time in a step and project it to probability simplex.

Although for every instance $\boldsymbol{d}_{i}$, in  \cite{WInLDL}, the projection onto a probability simplex problem is solved using an $O(m \log m)$ \cite{projection} algorithm by brute force searching through $[\mathrm{m}]$ from the largest entry to the smallest one for a particular $j$ satisfying the KKT condition, this projection method is not efficient. We use a much simpler projection function in $O(m)$:
\begin{equation}
\label{projection}
\begin{split}
    &\text{Step I: Setting all the } D_{ij} \text{ to } 0 \text{ if } D_{ij} < 0, \\
    &\text{Step II: Normalize $\mathbf{D}$ as }  D_{ij} = D_{ij} / \sum_{k = 1}^{m} D_{ik}.
\end{split}
\end{equation}
We set a uniform distribution if $\sum\limits_{k=1}^{m} D_{ik}= 0$. This projection method can be realized in $O(m)$, which is much faster than  \cite{projection}. At the same time, it can maintain proportional information and ultimately achieve better results in experiments.

2) $\mathbf{A}$ Subproblem: Removing irrelated terms regarding $\mathbf{A}$, $\mathbf{A}$ subproblem can be 
rewritten into

\begin{equation}
\label{A-op}
\min _{\mathbf{A}} \ \frac{1}{2} \| \mathbf{A}-(\mathbf{D}+\frac{\mathbf{\Lambda}}{\rho} )\|_F^{2}+\frac{\alpha}{\rho} \|\mathbf{A}\|_*,
\end{equation}
which has a closed-form solution by using a singular
value thresholding operator  \cite{singular}.

3) $\mathbf{B}$ Subproblem: Removing irrelated terms regarding $\mathbf{B}$, $\mathbf{B}$ subproblem can be 
solved by optimizing each
row of $\mathbf{B}$,

\begin{equation}
\label{op-B}
\begin{aligned}
&\min _{\boldsymbol{b}_i}\ \langle\boldsymbol{\lambda}^{'}_{i}, \boldsymbol{d}_i-\boldsymbol{b}_i\rangle+\frac{\rho}{2}\|\boldsymbol{d}_i-\boldsymbol{b}_i\|_F^2 \\
& \text{ s.t. } \boldsymbol{b}_i \in \left\{\boldsymbol{e}_i \mid \mathbf{E} \in Cons\right\},
\end{aligned}
\end{equation}
where the $i$-th row of $\mathbf{B}$ is denoted as $\boldsymbol{b}_i$, so is $\boldsymbol{\lambda}^{'}_{i}$, $\boldsymbol{d}_i$, $\boldsymbol{m}_i$, and $\boldsymbol{e}_i$. Note that in Eq. (\ref{op-B}), if ${M}_{ij} = 0$, no matter what $k_{i}$ is, $k_i \cdot D^{o}_{ij}=B_{ij} \cdot 
{M}_{ij} =0$. As a result, for those hidden positions, constraint is automatically satisfied. This problem turns out to be an unconstrained quadratic optimization, and $B_{ij}$ is directly set as $D_{i j}+\frac{\Lambda^{'}_{ij}}{\rho}$. When it comes to observed positions, $B_{ij} = k_i \cdot D^{o}_{ij}$, where $D^{o}_{ij}$ is a constant. Replacing 
all the $B_{ij}$ with $k_{i}$, the optimization object  is a quadratic equation about $k_{i}$, which is easy to deal with. Therefore, $\mathbf{B}$ can be updated as follows: 
\begin{equation}
\label{B-op}
B_{i j}= \begin{cases}
D_{i j} + \dfrac{\Lambda'_{ i j}}{\rho} &  M_{i j} = 0, \\
\dfrac{\sum_{k=1}^m \left(\rho D_{i k} + \Lambda'_{ i k}\right) \cdot D^o_{ik} \cdot M_{i k}}{\rho \sum_{k=1}^m(D^o_{ik})^2 \cdot M_{i k}} D^o_{ij} &  M_{ij} = 1.
\end{cases}
\end{equation}
\vspace{0.3cm}
Note that at least one entry
in $\boldsymbol{d}^{o}_i$ is not 0, so $\rho \sum_{k=1}^m( D^{o}_{ik}) ^{2} \cdot M_{ik}> 0$. As a result, Eq. (\ref{B-op}) is always meaningful and $B_{ij}$ can be updated correctly.

Furthermore, we update the Lagrange multiplier matrices as follows: 
\begin{equation}
\label{l-op}
\left\{\begin{array}{l}
\mathbf{\Lambda }=\mathbf{\Lambda }+\rho(\mathbf{D}-\mathbf{A}), \\
\mathbf{\Lambda}'=\mathbf{\Lambda}'+\rho(\mathbf{D}-\mathbf{B}).
\end{array}\right.
\end{equation}

See the overall pseudocode of the solution to Eq. (\ref{final_object}) in Algorithm \ref{alg:algorithm1}. Besides, the stopping criterion of the algorithm is that the number of iteration reaches 100, or the maximum of all the residuals of the optimized variables is less than $10^{-3}$, i.e., $\max \left(\|\mathbf{D}-\mathbf{A}\|_{\infty}\right.$, $\left.\|\mathbf{D}-\mathbf{B}\|_{\infty}\right)<10^{-3}$, where $\|\cdot\|_{\infty}$ denotes the infinity norm of the matrix. The  convergence of Eq. (\ref{final_object}) is theoretically guaranteed and we provide a proof in
Section \textbf{E} of the Appendix.

\begin{algorithm}[tb]
    \caption{Optimization to Eq. (\ref{final_object})}
    \label{alg:algorithm1}
    \setcounter{ALC@line}{0} 
    
    \begin{algorithmic}[1] 
        \STATE \textbf{Input}: $\mathbf{D}_o, \text{ the Laplacian matrix } \mathbf{G}, \text{ mask } \mathbf{M}$\\
        \STATE \textbf{Initialization}: $\mathbf{A} = \mathbf{B} = \mathbf{\Lambda}_1 = \mathbf{\Lambda}_2 = \mathbf{1}_{n \times m}, \mathbf{D} = \mathbf{D}_o,\ t = 1,
        \rho = 2
        $\WHILE{stopping criterion is not satisfied}
        \STATE Update $\mathbf{D}$ by Eq. (\ref{SGD});
        \STATE Project $\mathbf{D}$ by Eq. (\ref{projection});
        \STATE Solve $\mathbf{A}$ by Eq. (\ref{A-op});
        \STATE Update $\mathbf{B}$ by Eq. (\ref{B-op});
        \STATE Update $\mathbf{\Lambda, \Lambda'}$ by Eq. (\ref{l-op});
        \STATE $t = t + 1$;
        \ENDWHILE
        \STATE \textbf{return} $\mathbf{D}$
    \end{algorithmic}
\end{algorithm}

\section{Experiment}

\subsection{Datasets}


In this paper, we evaluate our method on 12 real-world datasets covering fields of biology, facial expression, natural scene, emotion recognition and movie.  The statistics of 12 datasets are provided in Section $\textbf{A}$ of the Appendix, and the details of them can be found in \cite{LDL,emotion6,RAF_ML}.

To simulate the presence of hidden labels in the dataset, we will first make these datasets random missing \cite{inLDL} and set the missing $d_{\boldsymbol{x}}^{y}$ to 0. We vary the missing rate $\omega$ from 40\% to 80\%. Then, we check the missing datasets and ensure that for each instance $\boldsymbol{x}$, at least one label description has a value greater than 0 (i.e., $\exists j$ such that $d_{\boldsymbol{x}}^{y_{j}} > 0$). In HidLDL, the input is a seemingly complete label distribution dataset, so in the second step, we normalize missing datasets into the final hidden setting using Eq. (\ref{normalize}). 

\begin{table*}[ht]
    \small
    \centering
    \setlength{\tabcolsep}{0.6mm}
    \resizebox{0.95\textwidth}{!}{
    \begin{tabular}{cccccccc}\hline
        Dataset & Ours & InLDL-a & WInLDL & SA-IIS & LDL-LRR & PT-Bayes & LDL-DPA  \\ \hline
        alpha & $\mathbf{0.4718 \pm .0011}$ & $\underline{0.6878 \pm .0010}^{\bullet}$ & $0.9660 \pm .0179^{\bullet}$ & $0.9466 \pm .0084^{\bullet}$ & $0.7078 \pm .0056^{\bullet}$ & $1.9636 \pm .0673^{\bullet}$ & $1.0037 \pm .0117^{\bullet}$  \\ 
        cdc & $\mathbf{0.3803 \pm .0044}$ & $\underline{0.6493 \pm .0023}^{\bullet}$ & $0.8870 \pm .0172^{\bullet}$ & $0.8590 \pm .0154^{\bullet}$ & $0.6607 \pm .0050^{\bullet}$ & $1.6762 \pm .0708^{\bullet}$ & $0.9007 \pm .0038^{\bullet}$  \\ 
        cold & $\mathbf{0.1746 \pm .0005}$ & $\underline{0.2442} \pm .0013 ^{\bullet}$& $0.2800 \pm .0049^{\bullet}$ & $0.2918 \pm .0045^{\bullet}$ & $0.2519 \pm .0045^{\bullet}$ & $0.4631 \pm .0551^{\bullet}$ & $0.3068 \pm .0086^{\bullet}$  \\ 
        dtt & $\mathbf{0.1237 \pm .0008}$ & $\underline{0.1734 \pm .0008}^{\bullet}$ & $0.2200 \pm .0058^{\bullet}$ & $0.2328 \pm .0063^{\bullet}$ & $0.1825 \pm .0062^{\bullet}$ & $0.4363 \pm .0410^{\bullet}$ & $0.2499 \pm .0032^{\bullet}$  \\ 
        elu & $\mathbf{0.3458 \pm .0072}$ & $\underline{0.5899 \pm .0011}^{\bullet}$ & $0.8038 \pm .0166^{\bullet}$ & $0.7791 \pm .0151^{\bullet}$ & $0.6029 \pm .0073^{\bullet}$ & $1.4970 \pm .0896^{\bullet}$ & $0.8156 \pm .0223^{\bullet}$  \\ 
        spo & $\mathbf{0.3250 \pm .0064}$ & $\underline{0.5162 \pm .0011}^{\bullet}$ & $0.5619 \pm .0042^{\bullet}$ & $0.5646 \pm .0075^{\bullet}$ & $0.5202 \pm .0020^{\bullet}$ & $0.8059 \pm .0204^{\bullet}$ & $0.5886 \pm .0072^{\bullet}$  \\ 
        SJAFFE & $\mathbf{0.5983 \pm .0235}$ & $0.8592 \pm .0535^{\bullet}$ & $2.0855 \pm .1299^{\bullet}$ & $\underline{0.8421 \pm .0339}^{\bullet}$ & $0.8891 \pm .0050^{\bullet}$ & $0.9838 \pm .0659^{\bullet}$ & $2.0990 \pm .1017^{\bullet}$  \\ 
        Scene & $\mathbf{6.4190 \pm .0026}$ & $\underline{6.7305 \pm .0113}^{\bullet}$ & $6.7673 \pm .0186^{\bullet}$ & $6.7141 \pm .0078^{\bullet}$ & $6.7761 \pm .0050^{\bullet}$ & $7.2195 \pm .1133^{\bullet}$ & $6.7538 \pm .0139^{\bullet}$  \\ 
        Movie & $\mathbf{0.7494 \pm .0041}$ & $\underline{1.1138 \pm .0047}^{\bullet}$ & $1.5030 \pm .0083^{\bullet}$ & $1.6548 \pm .0117^{\bullet}$ & $1.1415 \pm .0068^{\bullet}$ & $4.2956 \pm .1179^{\bullet}$ & $1.4271 \pm .0389^{\bullet}$  \\ 
        SBU & $\mathbf{0.7028 \pm .0086}$ & $0.8375 \pm .0048^{\bullet}$ & $\underline{0.8278 \pm .0103}^{\bullet}$ & $0.8686 \pm .0041^{\bullet}$ & $0.8998 \pm .0036^{\bullet}$ & $0.8737 \pm .0088^{\bullet}$ & $0.8552 \pm .0062^{\bullet}$  \\ 
        Emo & $\mathbf{2.4372 \pm .1010}$ & $\underline{3.7105 \pm .0233}^{\bullet}$ & $3.7991 \pm .0087^{\bullet}$ & $3.7944 \pm .0125^{\bullet}$ & $3.7732 \pm .0046^{\bullet}$ & $3.8714 \pm .0415^{\bullet}$ & $3.7921 \pm .0150^{\bullet}$  \\ 
        RAF & $3.0345 \pm .0095$ & $4.6553 \pm .0079^{\bullet}$ & $5.2983 \pm .0065^{\bullet}$ & $\mathbf{2.9245 \pm .0096}^{\circ}$ & $2.9998 \pm .0054^{\circ}$ & $3.1821 \pm .0440^{\bullet}$ & $\underline{2.9374 \pm .0112}^{\circ}$  \\ \hline
    \end{tabular}}
    \caption{Canberra (the lower the better) results for the \textbf{recovery} setting on all datasets when missing rate $\omega$ = 50\%. The value is shown in mean±std form. Bold and underlined indicate the best and second best results, respectively. Under single-tailed paired t-test at a significance level
of 0.05, $\bullet$ means our method’s performance is statistically significantly better than the compared
method and $\circ$ means the compared
method's performance is statistically significantly better than ours.}
    \label{canberra-50-recovery}
    \vspace{-4mm}
\end{table*}


\subsection{Settings and Compared Methods}

To validate the effectiveness of our method, we design two experimental configurations. In the first setting, we use a hidden dataset and features of instances together to restore the true label distribution. Difference between the ground-truth and the recovered distribution matrix will be measured. In the second setting, to further verify the quality of the recovered label distribution, we train each recovered label distribution using the same simple LDL algorithm, such as SA-BFGS \cite{LDL}. We then use each trained model for prediction. The training and testing sets are partitioned with ratios of 0.8 and 0.2, respectively. We repeat each experiment 5 times and report the average results.


We compare our method with six methods. Two of them are state-of-the-art methods designed for IncomLDL including IncomLDL-admm (abbreviated as InLDL-a) \cite{inLDL} and WInLDL \cite{WInLDL}, which can still be used for the hidden setting. 
 Two state-of-the-art LDL methods named LDL-LRR \cite{LDL-LRR} and LDL-DPA \cite{LDL-DPA} are included. We also include two baselines named  SA-IIS and PT-Bayes \cite{LDL}. 
For comparsion methods, we use default parameters suggested in their original papers,  with the exception that we adjust the regularization parameter for IncomLDL-admm from $2^{\{-10,-9, \ldots, 9,10\}}$. For our method, the regularization parameter $\alpha$ is selected from $2^{\{-10,-9, \ldots, 9,10\}}$ based on the recover performance, while $\sigma$ and $\rho$ are set to be 1 and 2, respectively. In K-nearest neighbors algorithm, the number of neighbors is set as the number of labels.  Each method will face a matrix with exactly the same missing entries in each experiment. 


We use five metrics for the LDL with Hidden Labels problem, including Chebyshev, Clark, Canberra, Cosine and Intersection. Details of them can be found in \cite{LDL}. Among them, Canberra, Clark and Canberra measure the distance between two vectors, thus they are the lower the better. Cosine and Intersection measure the similarity between two vectors, thus they are the higher the better. 

\subsection{Recover Results and Discussions}
In the following, we report the recover results of different methods at different missing rate. Due to space limitation, here we only list representative results. The Canberra (the lower the better) results on all the dataset with $\omega = 50\%$ are shown in Table \ref{canberra-50-recovery}. Other results are shown in Section $\textbf{B.1}$ of the Appendix. Moreover, different from previous research, we also study impact of missing rates on the results, which are also presented in Section $\textbf{B.2}$ of the Appendix.

\begin{figure}[t]
    \centering
    \begin{subfigure}[b]{0.23\textwidth}
        \centering   \includegraphics[width=\textwidth]{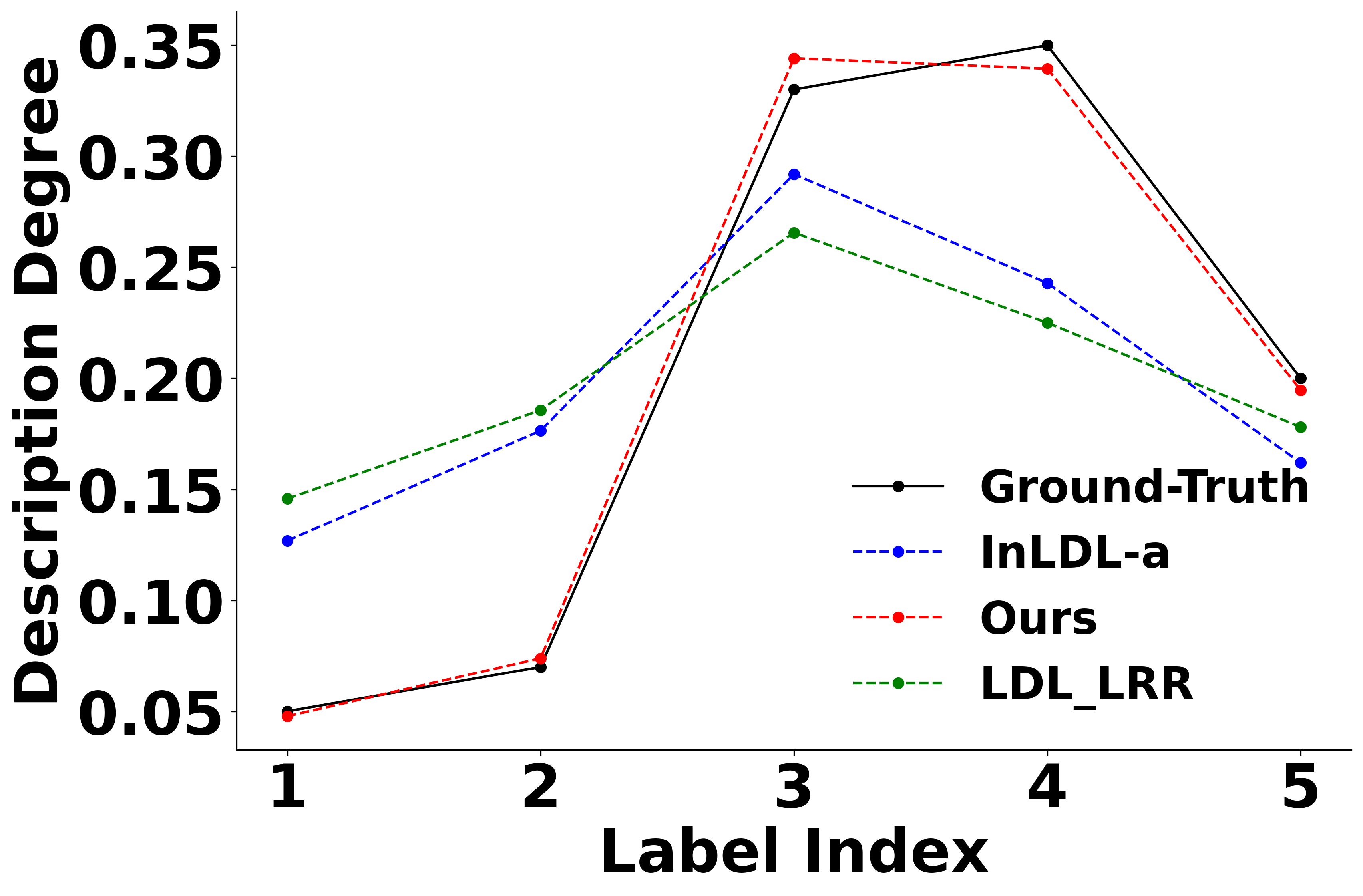}
        \caption{Movie}
    \end{subfigure}
    \hfill
    \begin{subfigure}[b]{0.23\textwidth}
        \centering      \includegraphics[width=\textwidth]{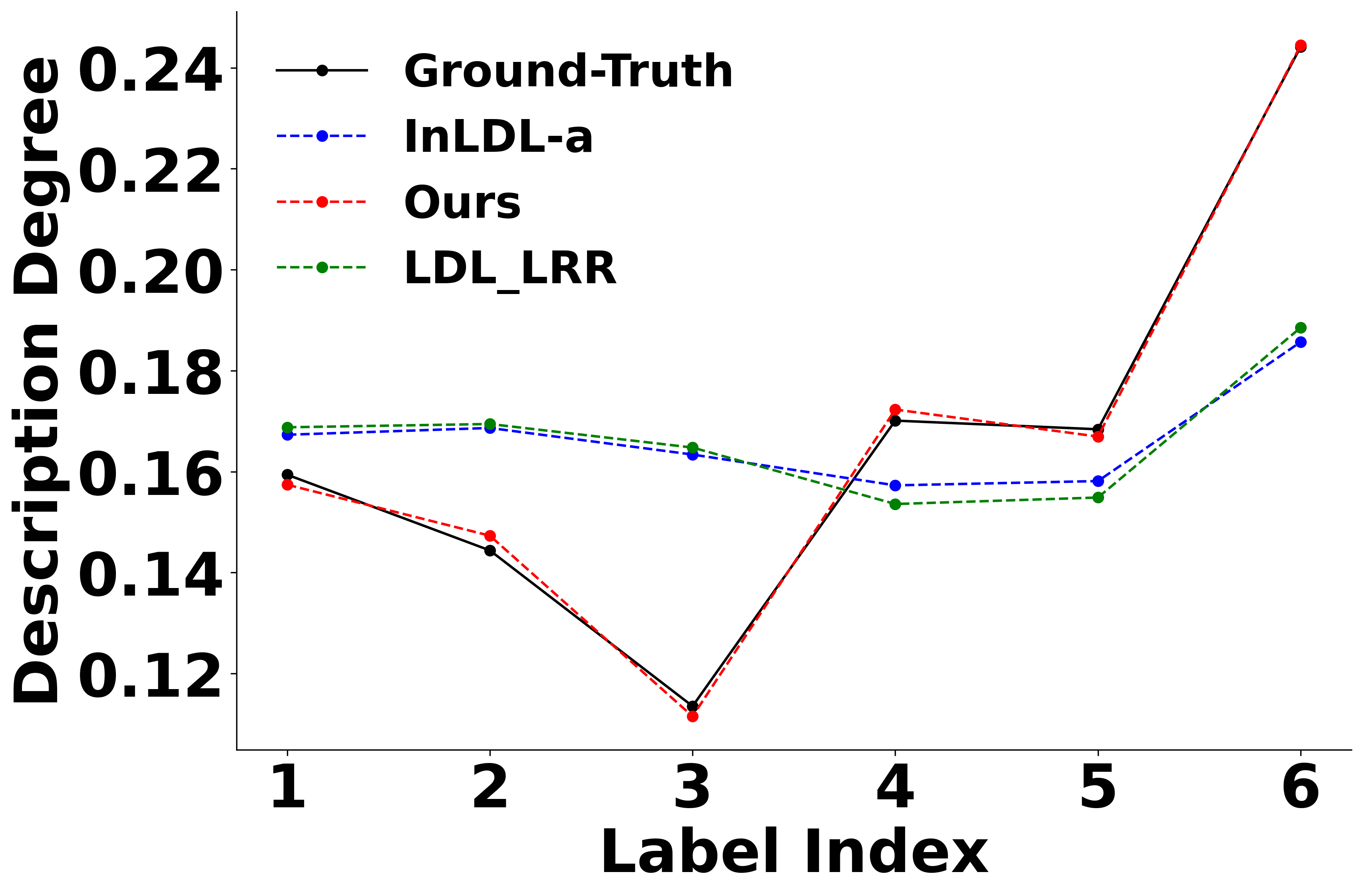}
        \caption{spo}
    \end{subfigure}
    \caption{The visualization of two typical recovery results on the
Movie (left) and spo (right) dataset.}
    \label{fig:visualization}
    \vspace{-2mm}
\end{figure}

To conduct significance analysis, we conduct the paired t-test, which is a useful statistical test for comparing two methods \cite{box1987guinness}. For each dataset, we perform 6 separate single-tailed paired t-tests, comparing our method with 6  compared methods individually. We mark the compared method with a black dot $\bullet$  in Table \ref{canberra-50-recovery} if our method's performance is statistically significantly better than the compared method at a significance level of 0.05.


By analyzing the experimental results, we can arrive at the following three conclusions: 

\begin{table}[t]
   \small
    \centering 
    \setlength{\tabcolsep}{1mm}
    \resizebox{0.46\textwidth}{!}{
    \begin{tabular}{ccccccccc}\hline
        Metric & Method & cold & dtt& spo & SBU & Emo & RAF \\
        \hline
        ~ & Ours & \textbf{0.099} & \textbf{0.072}&\textbf{0.163} & \textbf{0.327} & \textbf{1.249} & \textbf{1.486} \\ 
        Clark$\downarrow$ & w/o $Cons$ & 0.147 & 0.102&0.256 & 0.413 & 1.679 & 1.611 \\ 
        ~ & w/o TN & 0.123& 0.099 & 0.189 & 0.331 & 1.538 & 1.583 \\
        \hline
        ~ & Ours & \textbf{0.993} &\textbf{0.997}& \textbf{0.989} & \textbf{0.951} & \textbf{0.825} & \textbf{0.783} \\ 
        Cosine$\uparrow$& w/o $Cons$ & 0.987 &0.993& 0.975 & 0.918 & 0.657 & 0.642 \\ 
        ~ & w/o TN & 0.989 &0.996& 0.985 & 0.947 & 0.685 & 0.664 \\ 
        \hline
    \end{tabular}}
    \caption{
    Ablation Results on 6 Datasets. $\uparrow$ ($\downarrow$) indicates the higher (lower) the better.}
    \label{ablation_study}
    
    \vspace{-4mm}
\end{table}

\begin{table*}[!htbp]
    \small
    \centering
    \setlength{\tabcolsep}{0.6mm}
    \resizebox{0.95\textwidth}{!}{
    \begin{tabular}{cccccccc} \hline
        Dataset & Ours & InLDL-a & WInLDL & SA-IIS & LDL-LRR & PT-Bayes & LDL-DPA \\ \hline
        alpha & \textbf{0.6890 ± .0012}& $\underline{0.6936 \pm .0022}^{\bullet}$ & $1.0180 \pm .0094^{\bullet}$ & $0.9681 \pm .0138^{\bullet}$ & $0.7073 \pm .0101^{\bullet}$ & $1.8878 \pm .0996^{\bullet}$ & $1.0802 \pm .0173^{\bullet}$ \\ 
        cdc & \textbf{0.6464 ± .0020} & \underline{0.6506 ± .0056} & $0.9562 \pm .0250^{\bullet}$ & $0.8731 \pm .0175^{\bullet}$ & $0.6641 \pm .0095^{\bullet}$ & $1.7754 \pm .1194^{\bullet}$ & $0.9767 \pm .0178^{\bullet}$ \\ 
        cold & \textbf{0.2329 ± .0005} & $\underline{0.2369 \pm .0021}^{\bullet}$ & $0.2930 \pm .0117^{\bullet}$ & $0.2834 \pm .0071^{\bullet}$ & $0.2466 \pm .0038^{\bullet}$ & $0.5084 \pm .0815^{\bullet}$ & $0.3148 \pm .0128^{\bullet}$ \\ 
        dtt &\textbf{0.1635 ± .0002} & $\underline{0.1674 \pm .0010}^{\bullet}$ & $0.2370 \pm .0120^{\bullet}$ & $0.2351 \pm .0149^{\bullet}$ & $0.1916 \pm .0126^{\bullet}$ & $0.4543 \pm .0412^{\bullet}$ & $0.2714\pm .0215^{\bullet}$ \\ 
        elu & \textbf{0.5701 ± .0009} & $\underline{0.5771 \pm .0035}^{\bullet}$ & $0.8423 \pm .0178^{\bullet}$ & $0.7835 \pm .0119^{\bullet}$ & $0.5931 \pm .0057^{\bullet}$ & $1.5758 \pm .1034^{\bullet}$ & $0.8857 \pm .0186^{\bullet}$ \\ 
        spo & \textbf{0.5169 ± .0011} & \underline{0.5197 ± .0039} & $0.5956 \pm .0137^{\bullet}$ & $0.5702 \pm .0132^{\bullet}$ & $0.5243 \pm .0048^{\bullet}$& $0.8229 \pm .0539^{\bullet}$ & $0.6182 \pm .0193^{\bullet}$ \\ 
        SJAFFE & \textbf{0.7935 ± .0182 }& $\underline{0.8940 \pm .0263}^{\bullet}$ & $2.0713 \pm .0657^{\bullet}$ & $0.8980 \pm .0248^{\bullet}$ & $0.9120 \pm .0093^{\bullet}$ & $1.0189 \pm .0281^{\bullet}$ & $1.9060 \pm .1365^{\bullet}$ \\ 
        Scene & \textbf{6.8070 ± .0137} & $6.8531 \pm .0254^{\bullet}$ & $6.9508 \pm .0242^{\bullet}$ & \underline{6.8116 ± .0127} & $6.8298 \pm .0083^{\bullet}$ & $7.2288 \pm .0596^{\bullet}$ & $6.8700 \pm .0238^{\bullet}$ \\ 
        Movie & \textbf{1.0787 ± .0080} & $1.2025 \pm .0089^{\bullet}$ & $1.7374 \pm .0250^{\bullet}$ & $1.9103 \pm .0234^{\bullet}$ & $\underline{1.1610 \pm .0082}^{\bullet}$ &$ 4.2877 \pm .0859^{\bullet}$ & $1.4859 \pm .0210^{\bullet}$ \\ 
        SBU &\underline{0.8411 ± .0076} & $0.8716 \pm .0053^{\bullet}$ & $\mathbf{0.8328 \pm .0092}^{\circ}$ & $0.9017 \pm .0037^{\bullet}$ & $0.9142 \pm .0007^{\bullet}$ & $0.8988 \pm .0071^{\bullet}$ & $0.8892 \pm .0094^{\bullet}$ \\ 
        Emo & \textbf{3.7057 ± .0058} & $3.9040 \pm .0219^{\bullet}$ & $\underline{3.8321 \pm .0096}^{\bullet}$ & $4.2095 \pm .0456^{\bullet}$ & $3.8376 \pm .0373^{\bullet}$ & $3.8614 \pm .0815^{\bullet}$ & $3.9123 \pm .0411^{\bullet}$ \\ 
        RAF& 3.0544 ± .0043 & $\mathbf{2.8871 \pm .0133}^{\circ}$ & $4.3674 \pm .0360^{\bullet}$ & $3.0108 \pm  .0123^{\circ}$ &  $3.0033 \pm .0108^{\circ}$ & 3.0488 ± .0245 & $\underline{2.9869 \pm .0151}^{\circ}$ \\ \hline
    \end{tabular}}
    \caption{Canberra (the lower the better) results for the \textbf{predictive} setting on all datasets when missing rate $\omega$ = 50\%. The value is shown in mean±std form. Bold and underlined indicate the best and second best results, respectively. Under single-tailed paired t-test at a significance level
of 0.05, $\bullet$ means our method’s performance is statistically significantly better than the compared
method and $\circ$ means the compared
method's performance is statistically significantly better than ours.}
    \label{Canberra}
    \vspace{-3mm}
\end{table*}

\begin{itemize}
\item Our method achieves the lowest average rank in terms of all
five metrics. As shown in Table \ref{canberra-50-recovery}, we win 69 times out of 72 Canberra comparisons, with a \textbf{95.8\%} rate to win, demonstrating that our method is superior to other compared methods. More importantly, our advantages are statistically significant on these 69 cases according to t-test. For detailed metrics,
our improvement is particularly evident in Chebyshev, Clark and Canberra.

\item IncomLDL methods are not much better than LDL methods in the recovery setting. WInLDL, although reported to be faster and more effective in the incomplete setting  \cite{WInLDL}, performs quite poorly. InLDL-a, although performs relatively stably, is actually close to LDL-LRR in metrics' values. 
\item As the missing rate increases, the performance metrics of almost all methods become worse. Our method has an outstanding advantage when the missing rate is between 40\% and 60\%, as shown in Section \textbf{B.2} of the Appendix.
\end{itemize}

Fig. \ref{fig:visualization}
shows typical recovery results on two datasets. We selected two methods, InLDL-a and LDL-LRR, which are reported to have relatively better performance, to compare with our method.  According to the visualization, it can be seen that our method (\textbf{red line}) is closer to the ground truth (\textbf{black line}). Meanwhile, our trend has greater consistency with the ground truth, which is helpful to identify labels with the largest and smallest description degrees.

\begin{figure}[t]
    \centering
    
    \begin{subfigure}[b]{0.23\textwidth}
        \centering   \includegraphics[width=\textwidth]{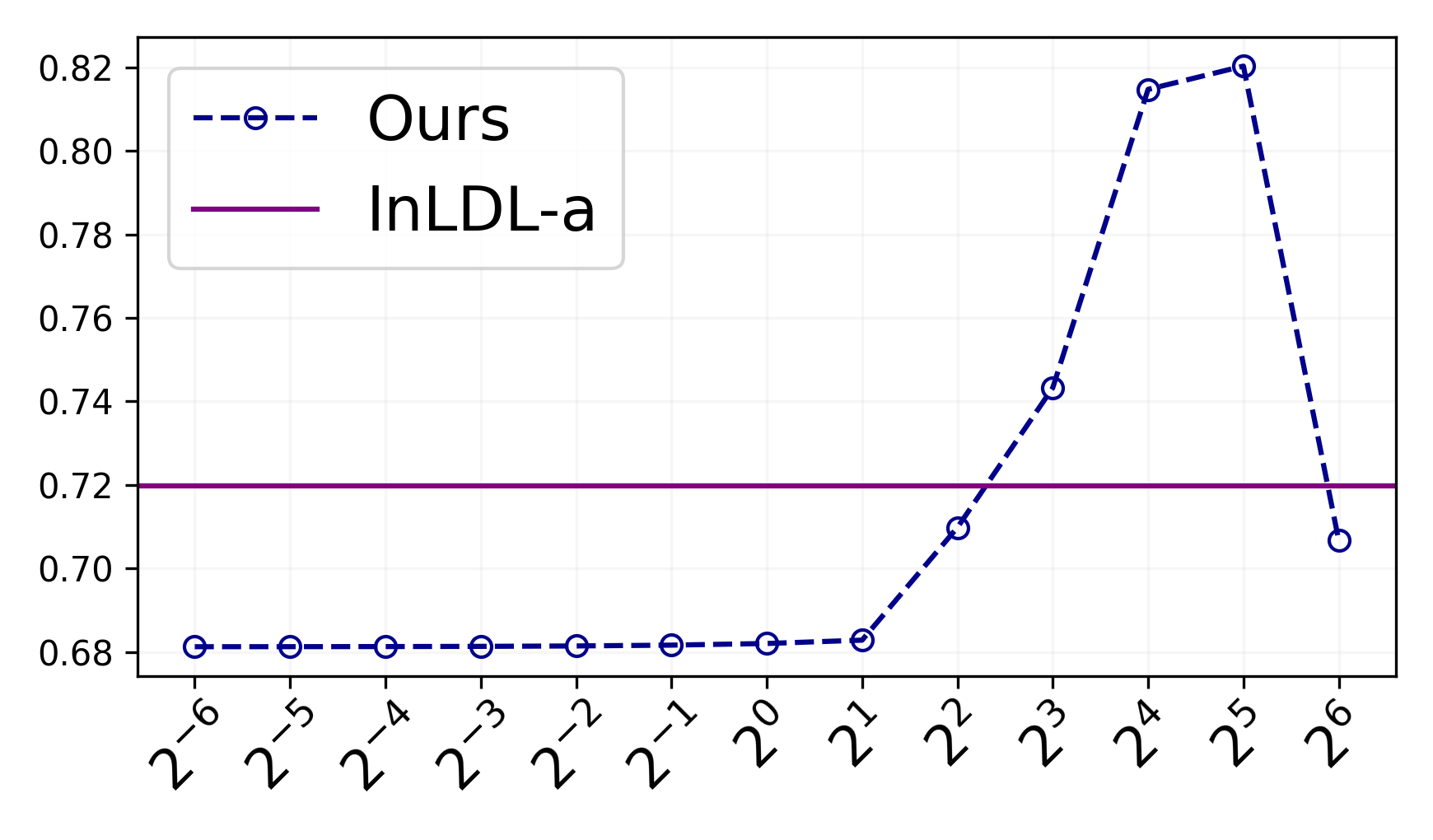}
        \caption{Emo}
        \label{Emotion6}
    \end{subfigure}
    \hfill
    \begin{subfigure}[b]{0.23\textwidth}
        \centering      \includegraphics[width=\textwidth]{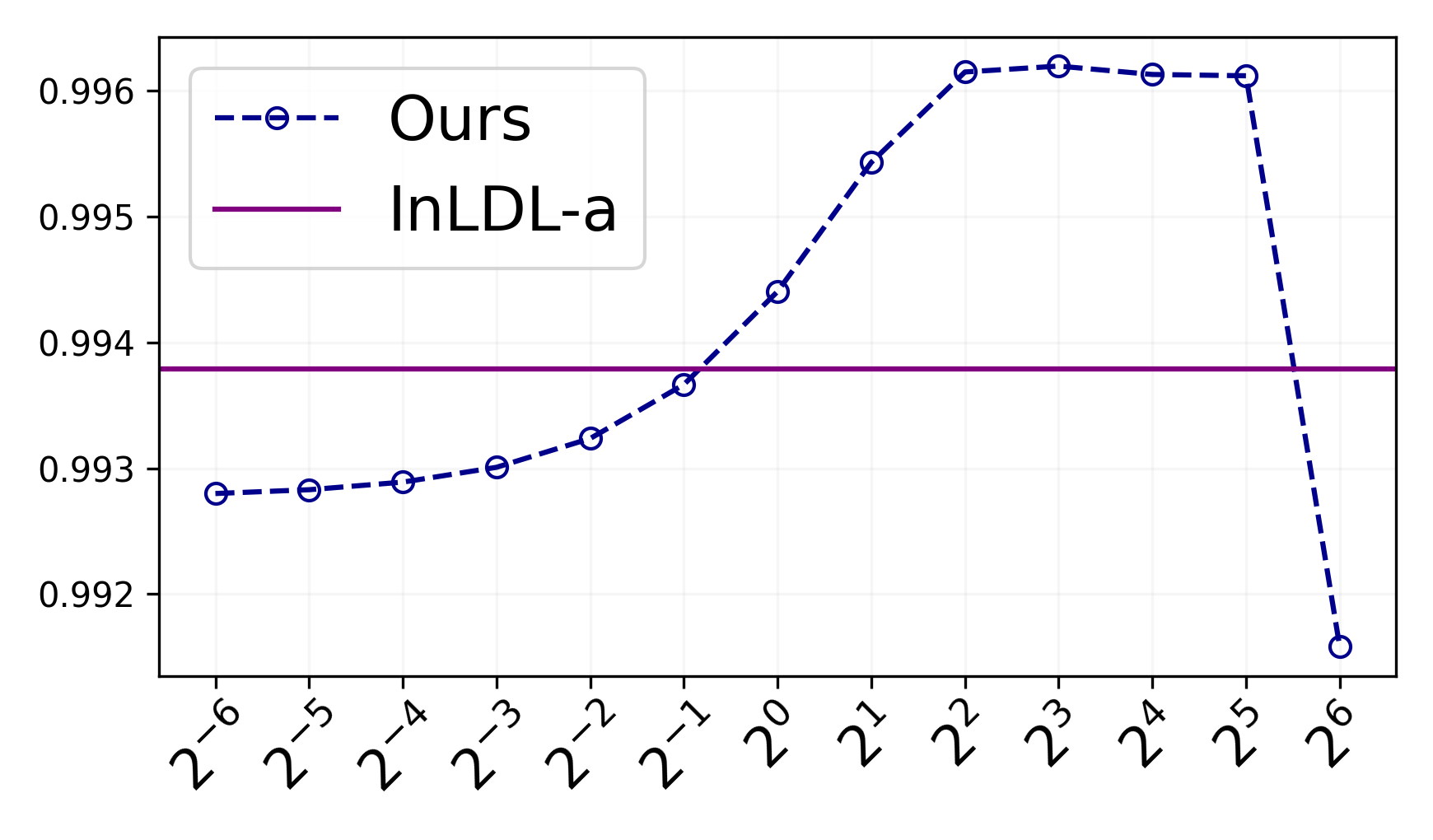}
        \caption{dtt}
        \label{Yeast-dtt}
    \end{subfigure}
    \hfill
    \begin{subfigure}[b]{0.23\textwidth}
        \centering     \includegraphics[width=\textwidth]{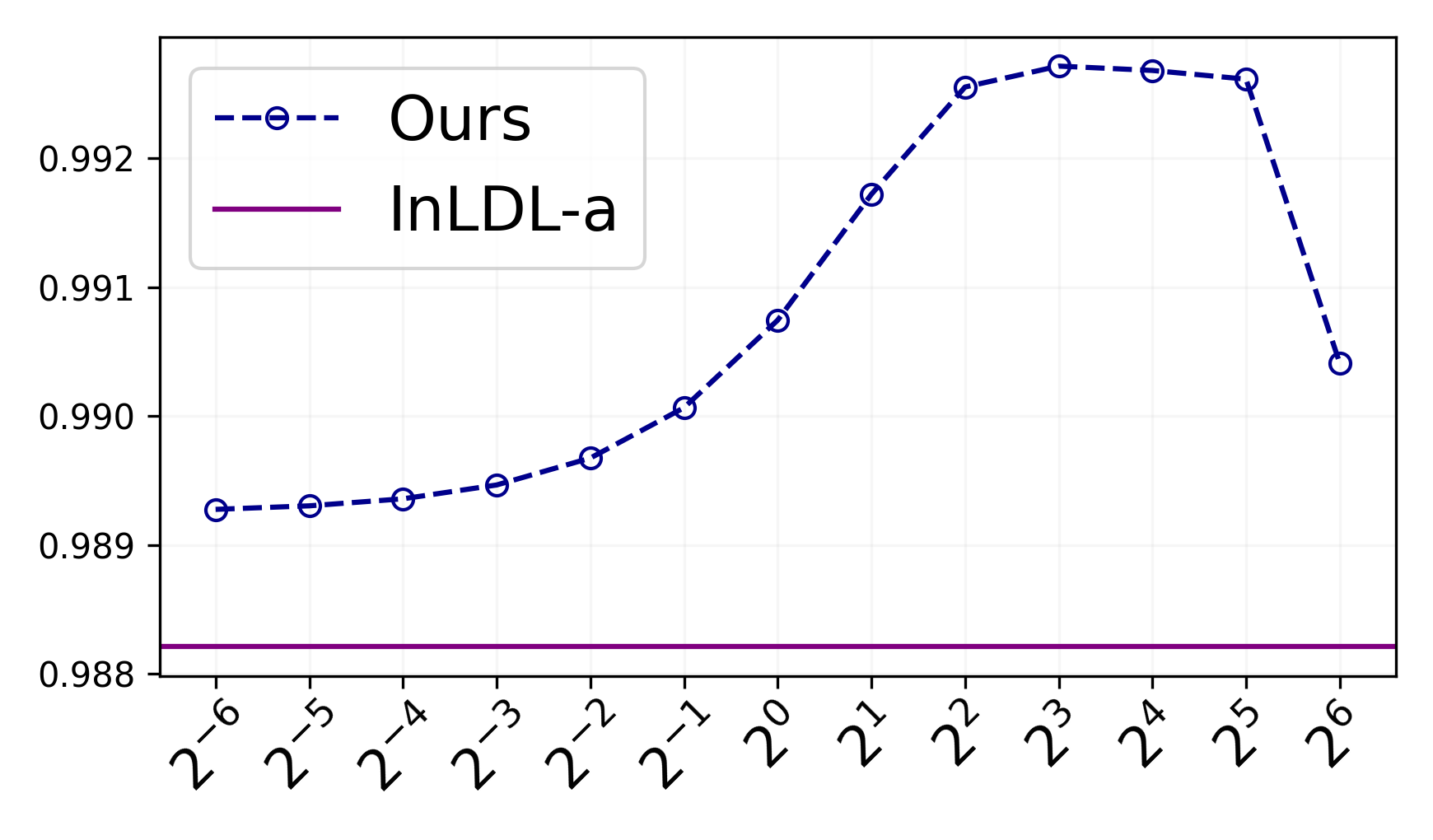}
        \caption{cold}
        \label{Yeast-cold}
    \end{subfigure}
    \hfill
    \begin{subfigure}[b]{0.23\textwidth}
        \centering        \includegraphics[width=\textwidth]{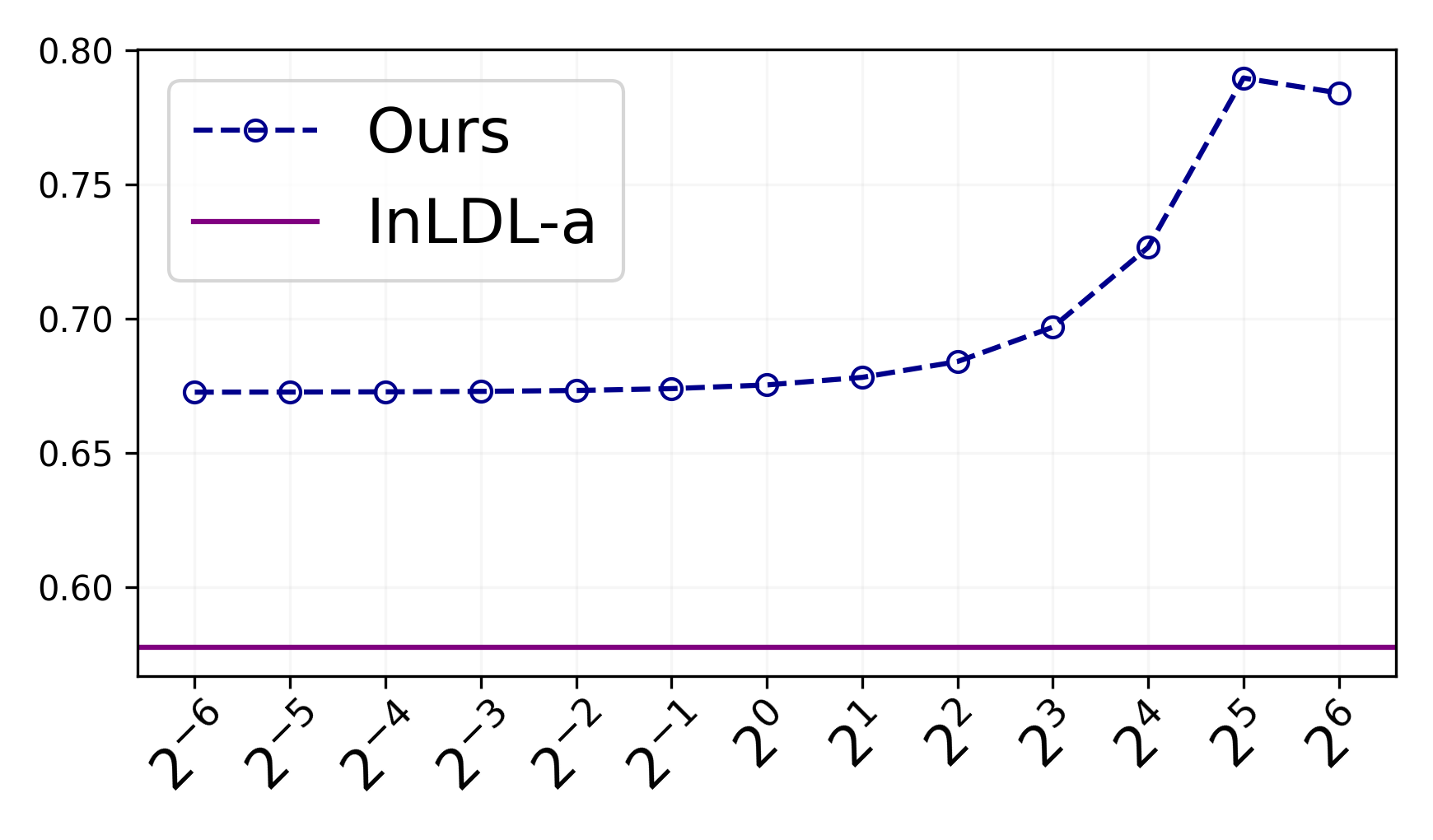}
        \caption{RAF}
        \label{RAF-sen}
    \end{subfigure}
    \caption{Comparison of our method and IncomLDL-admm on Cosine (the higher the better) and the horizontal axis represents the hyper-parameter $\alpha$ with the missing rate $\omega$=50\%.}
    \label{fig:images_hyper}
    \vspace{-4mm}
\end{figure}


\subsubsection{Ablation Studies}
We perform ablation experiments on six datasets to validate the importance of our proportional constraint and global dependency. As shown in Table \ref{ablation_study}, ``w/o $Cons$'' method means our method without proportional constraint, while ``w/o TN'' means our method without the trace norm (TN) term (i.e., $\alpha=0$). At 50\% missing rate, ``Ours'' has a huge improvement in all the evaluation metrics compared to the ``w/o $Cons$'', indicating that it is critical to maintain the proportional relationship between labels. Meanwhile, compared to ``w/o TN'', there has also been an improvement. Other results also show indispensability of $Cons$ and TN, which are provided in Section $\textbf{D}$ of the Appendix.

\subsubsection{Sensitivity Analysis}
Our method involves one parameter: $\alpha$ in the trace norm regularization. The analysis of the parameter $\alpha$ is shown in Fig. \ref{fig:images_hyper}. To demonstrate the performance of our method, we set the IncomLDL-admm method to achieve its best tuned results. From Fig. \ref{fig:images_hyper}, it can be seen that when $\alpha<2^3$, our method shows a stable growth in performance as $\alpha$ increases. When $\alpha>2^5$, the performance of our method deteriorates rapidly. As a result, when $\alpha$ is set between $2^3$ and $2^5$, relatively good results can be achieved.


\subsection{Predict Results and Discussions}
We present a representative result in Table \ref{Canberra} with a $50\%$ missing rate for the metric Canberra, while the results for other missing rates and metrics are similar, which are included in Section $\textbf{C}$ of the Appendix. Similar to the recovery setting, the single-tailed paired t-test is conducted to analyze the significant advantages of our method.
  

From the reported results, we observe that: 
\begin{itemize}
\item Our method achieves the lowest average rank in terms of all five metrics. Taking Table \ref{Canberra} as an example, out of 12 datasets, we rank 1st 10 times and 2nd once. Out of 72 comparisons, we win 66 times. With a winning rate of over \textbf{91.6\%}, our method performs better than most compared methods. And it is worth noting that our method has a significant advantage most of the time.

\item Methods focusing on the relationship between labels (such as LDL-LRR and InLDL-a) can capture advantageous information in prediction even if their recovery results are poor.
\item Our method is less affected by the missing rate in the predictive scenarios. As the missing rate increases, our method is robust and deteriorates very slowly. Relevant details are in Section $\textbf{C.2}$ of the Appendix.
\end{itemize}


\section{Conclusion}
This paper gives a creative answer to the question:   ``What does real-world incomplete data in IncomLDL actually look like and how to deal with it". Specifically, we clarify the irrationality of the training dataset in IncomLDL, and repair it by normalization. We then proposed a new method to solve this brand-new HidLDL problem, which emphasizes maintaining the observed labels' proportion during the recovery process. we then theoretically prove the feasibility of our method in learning from hidden labels. Furthermore, we theoretically provide the recovery bound of our method, demonstrating that it is feasible to learn from hidden labels. Extensive experiments validate advantages of our method against other IncomLDL and LDL methods in HidLDL.

\bibliography{aaai2026}

@article{liu2021emerging,
  title={The emerging trends of multi-label learning},
  author={Liu, Weiwei and Wang, Haobo and Shen, Xiaobo and Tsang, Ivor W},
  journal={IEEE transactions on pattern analysis and machine intelligence},
  volume={44},
  number={11},
  pages={7955--7974},
  year={2021},
  publisher={IEEE}
}

@article{LDL,
  author={Geng, Xin},
  journal={IEEE Transactions on Knowledge and Data Engineering}, 
  title={Label Distribution Learning}, 
  year={2016},
  volume={28},
  number={7},
  pages={1734-1748},
}

@article{MLL,
  title={Multi-label classification},
  author={Tsoumakas, Grigorios and Katakis, Ioannis},
  journal={Data Warehousing and Mining: Concepts, Methodologies, Tools, and Applications: Concepts, Methodologies, Tools, and Applications},
  volume={3},
  pages={64},
  year={2008},
  publisher={IGI Global}
}

@article{app1,
  title={Facial age estimation by learning from label distributions},
  author={Geng, Xin and Yin, Chao and Zhou, Zhi-Hua},
  journal={IEEE transactions on pattern analysis and machine intelligence},
  volume={35},
  number={10},
  pages={2401--2412},
  year={2013},
  publisher={IEEE}
}

@inproceedings{app2,
  title={Facial emotion distribution learning by exploiting low-rank label correlations locally},
  author={Jia, Xiuyi and Zheng, Xiang and Li, Weiwei and Zhang, Changqing and Li, Zechao},
  booktitle={Proceedings of the IEEE/CVF Conference on computer vision and pattern recognition},
  pages={9841--9850},
  year={2019}
}

@inproceedings{app3,
  title={SCUT-FBP5500: A diverse benchmark dataset for multi-paradigm facial beauty prediction},
  author={Liang, Lingyu and Lin, Luojun and Jin, Lianwen and Xie, Duorui and Li, Mengru},
  booktitle={2018 24th International conference on pattern recognition (ICPR)},
  pages={1598--1603},
  year={2018},
  organization={IEEE}
}

@inproceedings{app4,
  title={Pre-release prediction of crowd opinion on movies by label distribution learning.},
  author={Geng, Xin and Hou, Peng},
  booktitle={IJCAI},
  pages={3511--3517},
  year={2015},
  organization={Citeseer}
}

@article{app5,
title = {Multi-Class ASD Classification via Label Distribution Learning with Class-Shared and Class-Specific Decomposition},
journal = {Medical Image Analysis},
volume = {75},
pages = {102294},
year = {2022},
issn = {1361-8415},
doi = {https://doi.org/10.1016/j.media.2021.102294},
url = {https://www.sciencedirect.com/science/article/pii/S136184152100339X},
author = {Jun Wang and Fengyexin Zhang and Xiuyi Jia and Xin Wang and Han Zhang and Shihui Ying and Qian Wang and Jun Shi and Dinggang Shen},

}

@inproceedings{yang,
  title={Deep label distribution learning for apparent age estimation},
  author={Yang, Xu and Gao, Bin-Bin and Xing, Chao and Huo, Zeng-Wei and Wei, Xiu-Shen and Zhou, Ying and Wu, Jianxin and Geng, Xin},
  booktitle={Proceedings of the IEEE international conference on computer vision workshops},
  pages={102--108},
  year={2015}
}

@article{LDL-LDM,
  title={Label distribution learning by exploiting label distribution manifold},
  author={Wang, Jing and Geng, Xin},
  journal={IEEE transactions on neural networks and learning systems},
  volume={34},
  number={2},
  pages={839--852},
  year={2021},
  publisher={IEEE}
}

@article{LS,
  title={Label distribution learning with label correlations on local samples},
  author={Jia, Xiuyi and Li, Zechao and Zheng, Xiang and Li, Weiwei and Huang, Sheng-Jun},
  journal={IEEE Transactions on Knowledge and Data Engineering},
  volume={33},
  number={4},
  pages={1619--1631},
  year={2019},
  publisher={IEEE}
}

@inproceedings{LC,
  title={Label distribution learning by exploiting label correlations},
  author={Jia, Xiuyi and Li, Weiwei and Liu, Junyu and Zhang, Yu},
  booktitle={Proceedings of the AAAI Conference on Artificial Intelligence},
  volume={32},
  number={1},
  year={2018}
}

@inproceedings{inLDL,
  title={Incomplete Label Distribution Learning.},
  author={Xu, Miao and Zhou, Zhi-Hua},
  booktitle={IJCAI},
  pages={3175--3181},
  year={2017}
}

@article{inLDL-LCD,
  title={Incomplete label distribution learning via label correlation decomposition},
  author={Xu, Suping and Shang, Lin and Shen, Furao and Yang, Xibei and Pedrycz, Witold},
  journal={Information Fusion},
  volume={113},
  pages={102600},
  year={2025},
  publisher={Elsevier}
}

@inproceedings{WInLDL,
  title={No regularization is needed: efficient and effective incomplete label distribution learning},
  author={Li, Xiang and Chen, Songcan},
  booktitle={Proceedings of the Thirty-Third International Joint Conference on Artificial Intelligence},
  pages={4470--4478},
  year={2024}
}

@article{projection,
  title={Projection onto the probability simplex: An efficient algorithm with a simple proof, and an application},
  author={Wang, Weiran and Carreira-Perpin{\'a}n, Miguel A},
  journal={arXiv preprint arXiv:1309.1541},
  year={2013}
}

@article{singular,
  title={A singular value thresholding algorithm for matrix completion},
  author={Cai, Jian-Feng and Cand{\`e}s, Emmanuel J and Shen, Zuowei},
  journal={SIAM Journal on optimization},
  volume={20},
  number={4},
  pages={1956--1982},
  year={2010},
  publisher={SIAM}
}

@article{LDL-DPA,
  title={Adaptive weighted ranking-oriented label distribution learning},
  author={Jia, Xiuyi and Qin, Tian and Lu, Yunan and Li, Weiwei},
  journal={IEEE Transactions on Neural Networks and Learning Systems},
  year={2023},
  publisher={IEEE}
}

@article{LDL-LRR,
  title={Label distribution learning by maintaining label ranking relation},
  author={Jia, Xiuyi and Shen, Xiaoxia and Li, Weiwei and Lu, Yunan and Zhu, Jihua},
  journal={IEEE Transactions on Knowledge and Data Engineering},
  volume={35},
  number={2},
  pages={1695--1707},
  year={2021},
  publisher={IEEE}
}

@INPROCEEDINGS{ordinal,
  author={Wen, Changsong and Zhang, Xin and Yao, Xingxu and Yang, Jufeng},
  booktitle={2023 IEEE/CVF International Conference on Computer Vision (ICCV)}, 
  title={Ordinal Label Distribution Learning}, 
  year={2023},
  volume={},
  number={},
  pages={23424-23434},
  doi={10.1109/ICCV51070.2023.02146}}

@article{ADMM,
  title={Distributed optimization and statistical learning via the alternating direction method of multipliers},
  author={Boyd, Stephen and Parikh, Neal and Chu, Eric and Peleato, Borja and Eckstein, Jonathan and others},
  journal={Foundations and Trends{\textregistered} in Machine learning},
  volume={3},
  number={1},
  pages={1--122},
  year={2011},
  publisher={Now Publishers, Inc.}
}

@inproceedings{emotion6,
  title={A mixed bag of emotions: Model, predict, and transfer emotion distributions},
  author={Peng, Kuan-Chuan and Chen, Tsuhan and Sadovnik, Amir and Gallagher, Andrew C},
  booktitle={Proceedings of the IEEE conference on computer vision and pattern recognition},
  pages={860--868},
  year={2015}
}

@article{RAF_ML,
  title={Blended emotion in-the-wild: Multi-label facial expression recognition using crowdsourced annotations and deep locality feature learning},
  author={Li, Shan and Deng, Weihong},
  journal={International Journal of Computer Vision},
  volume={127},
  number={6},
  pages={884--906},
  year={2019},
  publisher={Springer}
}

@book{smooth,
  title={Semi-supervised learning with graphs},
  author={Zhu, Xiaojin},
  year={2005},
  publisher={Carnegie Mellon University}
}

@article{box1987guinness,
  title={Guinness, Gosset, Fisher, and small samples},
  author={Box, Joan Fisher},
  journal={Statistical science},
  pages={45--52},
  year={1987},
  publisher={JSTOR}
}

@article{MLL2,
  author    = {Min{-}Ling Zhang and
               Zhi{-}Hua Zhou},
  title     = {A Review on Multi-Label Learning Algorithms},
  journal   = {{IEEE} Transactions on Knowledge and Data Engineering},
  volume    = {26},
  number    = {8},
  pages     = {1819--1837},
  year      = {2014},
}

@article{app6,
  author       = {Haoyu Ma and
                  Ningning Lu and
                  Junjun Mei and
                  Tao Guan and
                  Yu Zhang and
                  Xin Geng},
  title        = {Label distribution learning for scene text detection},
  journal      = {Frontiers Comput. Sci.},
  volume       = {17},
  number       = {6},
  pages        = {176339},
  year         = {2023},
  url          = {https://doi.org/10.1007/s11704-022-1446-5},
}

@inproceedings{app1_2,
  author       = {Bin{-}Bin Gao and
                  Hong{-}Yu Zhou and
                  Jianxin Wu and
                  Xin Geng},
  editor       = {J{\'{e}}r{\^{o}}me Lang},
  title        = {Age Estimation Using Expectation of Label Distribution Learning},
  booktitle    = {Proceedings of the Twenty-Seventh International Joint Conference on
                  Artificial Intelligence, {IJCAI} 2018, July 13-19, 2018, Stockholm,
                  Sweden},
  pages        = {712--718},
  publisher    = {ijcai.org},
  year         = {2018},
}

@inproceedings{app2_2,
  author       = {Shikai Chen and
                  Jianfeng Wang and
                  Yuedong Chen and
                  Zhongchao Shi and
                  Xin Geng and
                  Yong Rui},
  title        = {Label Distribution Learning on Auxiliary Label Space Graphs for Facial
                  Expression Recognition},
  booktitle    = {2020 {IEEE/CVF} Conference on Computer Vision and Pattern Recognition,
                  {CVPR} 2020, Seattle, WA, USA, June 13-19, 2020},
  pages        = {13981--13990},
  publisher    = {Computer Vision Foundation / {IEEE}},
  year         = {2020},
}

@inproceedings{app6_2,
  author       = {Yi Ren and
                  Xin Geng},
  editor       = {Carles Sierra},
  title        = {Sense Beauty by Label Distribution Learning},
  booktitle    = {Proceedings of the Twenty-Sixth International Joint Conference on
                  Artificial Intelligence, {IJCAI} 2017, Melbourne, Australia, August
                  19-25, 2017},
  pages        = {2648--2654},
  publisher    = {ijcai.org},
  year         = {2017},
  url          = {https://doi.org/10.24963/ijcai.2017/369},
}

@inproceedings{rankmatch,
  author       = {Kouzhiqiang Yucheng Xie and
                  Jing Wang and
                  Yuheng Jia and
                  Boyu Shi and
                  Xin Geng},
  title        = {RankMatch: {A} Novel Approach to Semi-Supervised Label Distribution
                  Learning Leveraging Inter-label Correlations},
  booktitle={Proceedings of the AAAI Conference on Artificial Intelligence},
  year={2026}
}

@inproceedings{concentration,
  author       = {Jiawei Tang and
                  Yuheng Jia},
  title        = {Concentration Distribution Learning from Label Distributions},
  booktitle    = {Forty-second International Conference on Machine Learning, {ICML} 2025, Vancouver, Canada, July 13-19, 2025},
  year={2025}
}

@inproceedings{DLDL,
  author       = {Yuheng Jia and
                  Jiawei Tang and
                  Jiahao Jiang},
  title        = {Label Distribution Learning from Logical Label},
  booktitle    = {Proceedings of the Thirty-Third International Joint Conference on
                  Artificial Intelligence, {IJCAI} 2024, Jeju, South Korea, August 3-9,
                  2024},
  pages        = {4228--4236},
  year         = {2024}
}

@article{instancedepend,
  author       = {Zhiqiang Kou and
                  Jing Wang and
                  Yuheng Jia and
                  Biao Liu and
                  Xin Geng},
  title        = {Instance-Dependent Inaccurate Label Distribution Learning},
  journal      = {{IEEE} Trans. Neural Networks Learn. Syst.},
  volume       = {36},
  number       = {1},
  pages        = {1425--1437},
  year         = {2025},
}

@inproceedings{biased,
  author       = {Zhiqiang Kou and
                  Si Qin and
                  Hailin Wang and
                  Jing Wang and
                  Ming{-}Kun Xie and
                  Shuo Chen and
                  Yuheng Jia and
                  Tongliang Liu and
                  Masashi Sugiyama and
                  Xin Geng},
  title        = {Label Distribution Learning with Biased Annotations Assisted by Multi-Label
                  Learning},
  booktitle    = {Proceedings of the Thirty-Fourth International Joint Conference on
                  Artificial Intelligence, {IJCAI} 2025, Montreal, Canada, August 16-22,
                  2025},
  pages        = {5545--5553},
  year         = {2025},
}

@inproceedings{complementary,
  author       = {Yuhang Li and
                  Zhuying Li and
                  Yuheng Jia},
  title        = {Complementary Label Learning with Positive Label Guessing and Negative
                  Label Enhancement},
  booktitle    = {The Thirteenth International Conference on Learning Representations,
                  {ICLR} 2025, Singapore, April 24-28, 2025},
  publisher    = {OpenReview.net},
  year         = {2025},
  url          = {https://openreview.net/forum?id=LPRxGZ7Oax},
  bibsource    = {dblp computer science bibliography, https://dblp.org}
}

@article{ensemble,
  author       = {Yuheng Jia and
                  Sirui Tao and
                  Ran Wang and
                  Yongheng Wang},
  title        = {Ensemble Clustering via Co-Association Matrix Self-Enhancement},
  journal      = {{IEEE} Trans. Neural Networks Learn. Syst.},
  volume       = {35},
  number       = {8},
  pages        = {11168--11179},
  year         = {2024},
}

@article{semisupervise,
  author       = {Yuheng Jia and
                  Hui Liu and
                  Junhui Hou and
                  Sam Kwong and
                  Qingfu Zhang},
  title        = {Semisupervised Affinity Matrix Learning via Dual-Channel Information
                  Recovery},
  journal      = {{IEEE} Trans. Cybern.},
  volume       = {52},
  number       = {8},
  pages        = {7919--7930},
  year         = {2022},
}

\clearpage

\appendix
\setcounter{figure}{0}
\renewcommand{\thefigure}{A\arabic{figure}}
\setcounter{table}{0}
\renewcommand{\thetable}{A\arabic{table}}

\section*{Appendix}

\section{Statistics of Datasets}

\begin{table}[H]

\centering
\begin{tabular}{cccc}
\hline
Dataset & $n$ & $d$ & $m$ \\
\hline
Yeast-alpha & 2465 & 24 & 18 \\
Yeast-cdc & 2465 & 24 & 15 \\
Yeast-cold & 2465 & 24 & 4 \\
Yeast-dtt & 2465 & 24 & 4 \\
Yeast-elu & 2465 & 24 & 14 \\
Yeast-spo & 2465 & 24 & 6 \\
SJAFFE & 213 & 243 & 6 \\
Natural Scene & 2000 & 294 & 9 \\
Movie & 7755 & 1869 & 5 \\
SBU3DFE & 2500 & 243 & 6 \\
Emotion6 & 1980 & 1000 & 7 \\
RAF-ML & 4908 & 200 & 6 \\
\hline
\end{tabular}
\caption{Statistics of the 12 datasets, where $n$, $d$ and $m$ represent the number of samples, the dimension of features, and the number of labels.}
\label{dataset}
\end{table}

Table \ref{dataset} shows the
statistics of 12 datasets. Due to space limitations, we only use the abbreviations of the datasets in the paper. The following are the abbreviations of each dataset.
\begin{itemize}
\item Yeast-alpha $\rightarrow$ alpha
\item Yeast-cdc $\rightarrow$ cdc
\item Yeast-cold $\rightarrow$ cold
\item Yeast-dtt $\rightarrow$ dtt
\item Yeast-elu $\rightarrow$ elu
\item Yeast-spo $\rightarrow$ spo
\item SJAFFE $\rightarrow$ SJAFFE
\item Natural Scene$\rightarrow$ Scene
\item Movie $\rightarrow$ Movie
\item SBU3DFE $\rightarrow$ SBU
\item Emotion6 $\rightarrow$ Emo
\item RAF-ML $\rightarrow$ RAF
\end{itemize}


\section{Recovery Results}
\subsubsection{More Recovery Results}
Table~\ref{Chebyshev} and Table~\ref{clark-80-recovery} show the recovery results for the evaluation metric Chebyshev at missing rate $\omega$ = 50\% and the evaluation metric Clark at missing rate $\omega$ = 80\%, respectively. As shown in table~\ref{Chebyshev} and table~\ref{clark-80-recovery}, our method outperform comparison methods by a wide margin. Specifically, in Table~\ref{Chebyshev}, our method ranks first 11 times out of 12 datasets. Similarly, in Table~\ref{clark-80-recovery}, it ranks first 10 times out of 12.
\subsection{Recovery Results ‌Versus Missing Rates}
\begin{figure*}[ht]
    \centering
    \begin{subfigure}[b]{0.45\textwidth} 
        \centering
        \includegraphics[width=\textwidth]{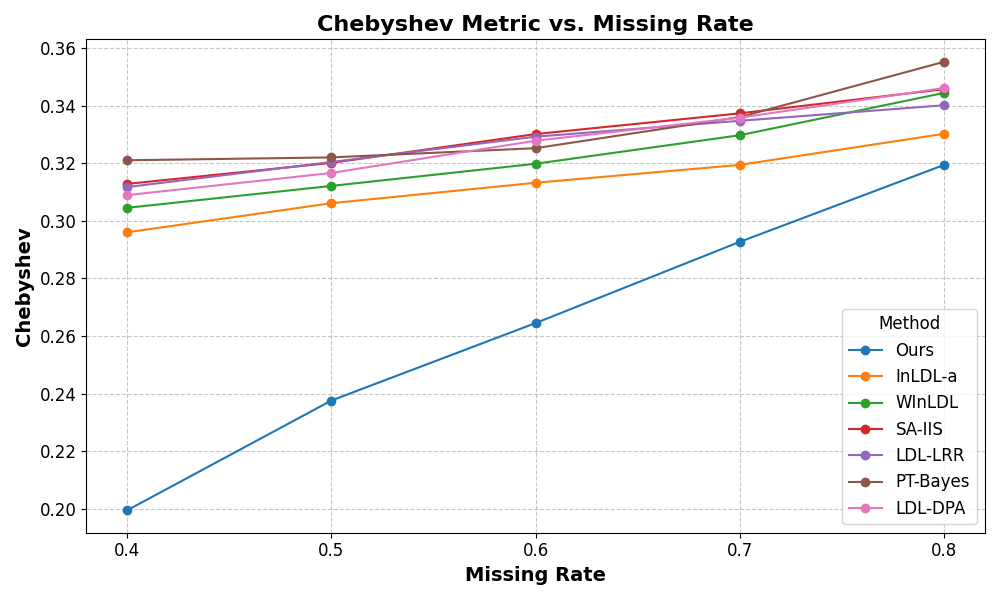}
        \caption{Emotion6}
        \label{Emotion6}
    \end{subfigure}
    \hspace{0.05\textwidth} 
    \begin{subfigure}[b]{0.45\textwidth} 
        \centering
        \includegraphics[width=\textwidth]{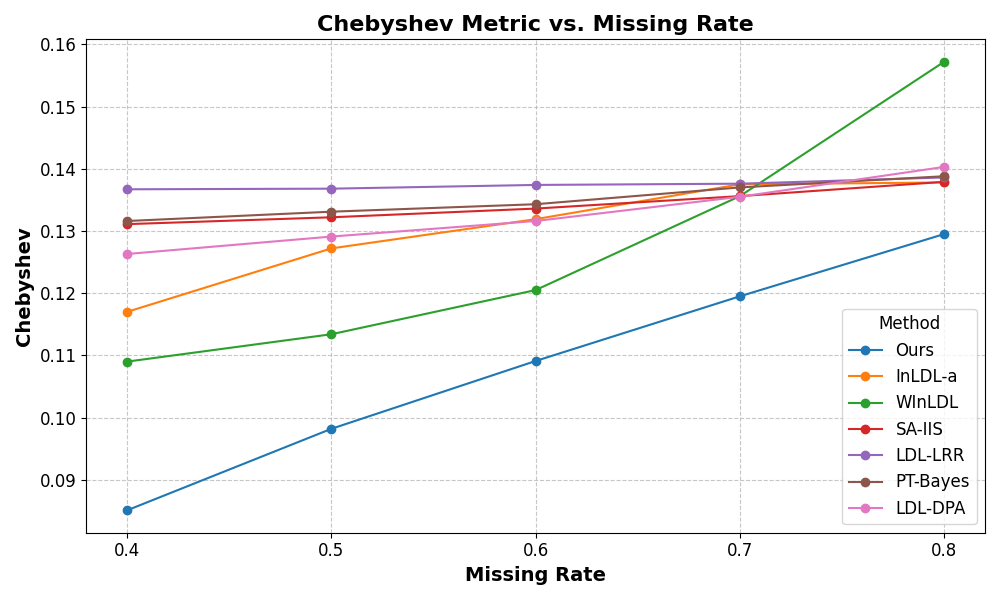}
        \caption{SBU3DFE}
        \label{SBU_3DFE}
    \end{subfigure}
    
    \vspace{0.2cm} 
    
    \begin{subfigure}[b]{0.45\textwidth} 
        \centering
        \includegraphics[width=\textwidth]{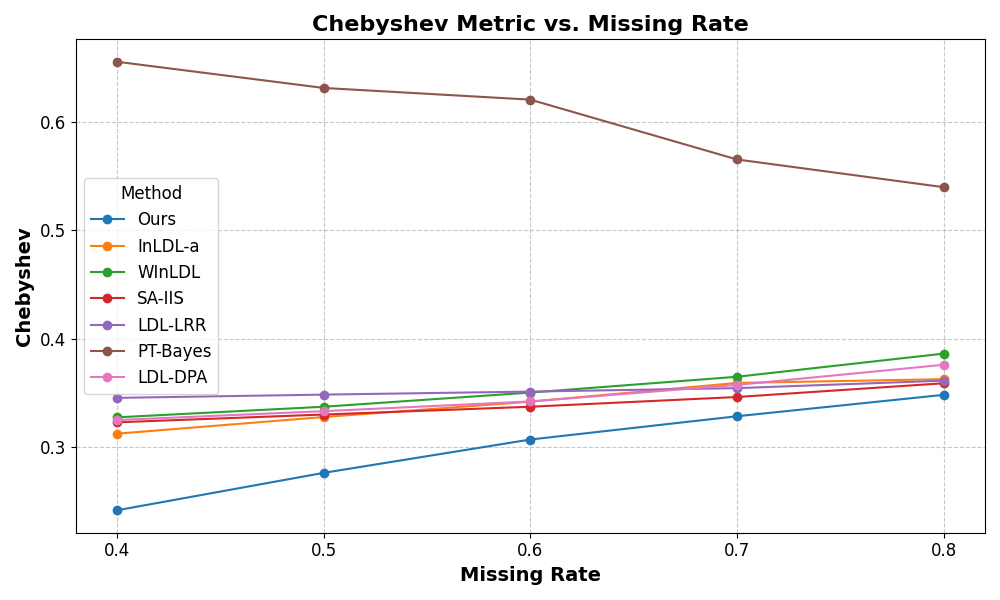}
        \caption{Natural Scene}
        \label{Scene}
    \end{subfigure}
    \hspace{0.05\textwidth} 
    \begin{subfigure}[b]{0.45\textwidth} 
        \centering
        \includegraphics[width=\textwidth]{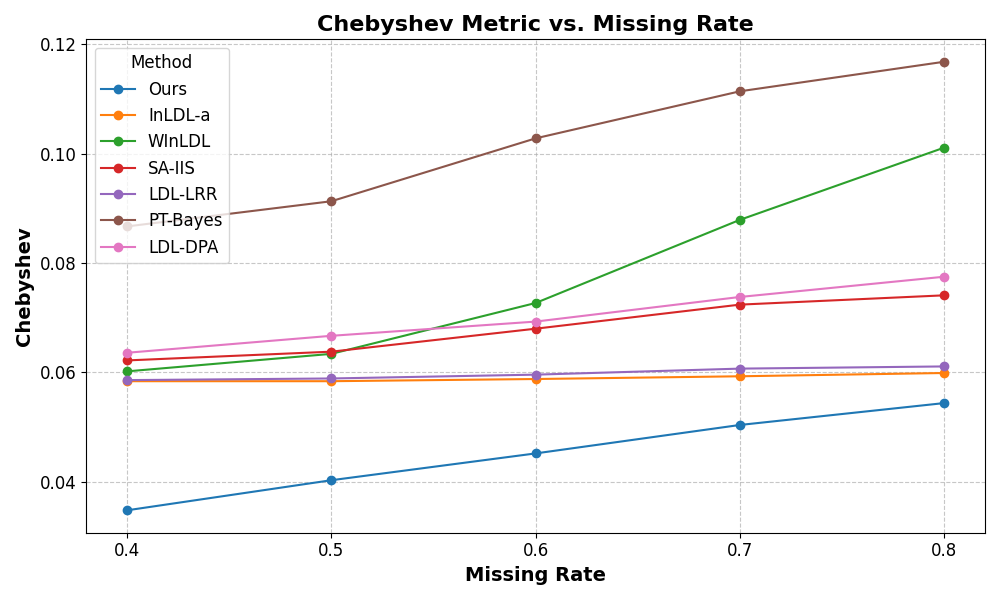}
        \caption{Yeast-spo}
        \label{Yeast_spo}
    \end{subfigure}
    
    \caption{Recovery performance comparison under different missing rates.}
    \label{fig:images_MRandPer}
\end{figure*}

Fig. \ref{fig:images_MRandPer} shows the results of all methods in the recovery experiment under different missing rates. In all of the 4 datasets, our method achieves better performance under different missing rates, which clearly indicate that our method is superior to the compared methods. Furthermore, as the missing rate decreases, the superiority of our method become more apparent. Our method has an outstanding advantage when the missing rate is between 40\% and 60\%.

\section{Predictive Results}

\subsection{More Predictive Results}

Table~\ref{cheby-50-predictive} and Table~\ref{clark-80-predictive} show the predictive results for the evaluation metrics of Chebyshev at missing rate $\omega$ = 50\% and Clark at missing rate $\omega$ = 80\%, respectively. In these two tables, our method outperforms comparison methods 21 out of 24 times, resulting in a leading rate of $87.5 \%$.

\subsection{Predictive Results ‌Versus Missing Rates}
\begin{figure*}[ht]
    \centering
    \begin{subfigure}[b]{0.45\textwidth} 
        \centering
        \includegraphics[width=\textwidth]{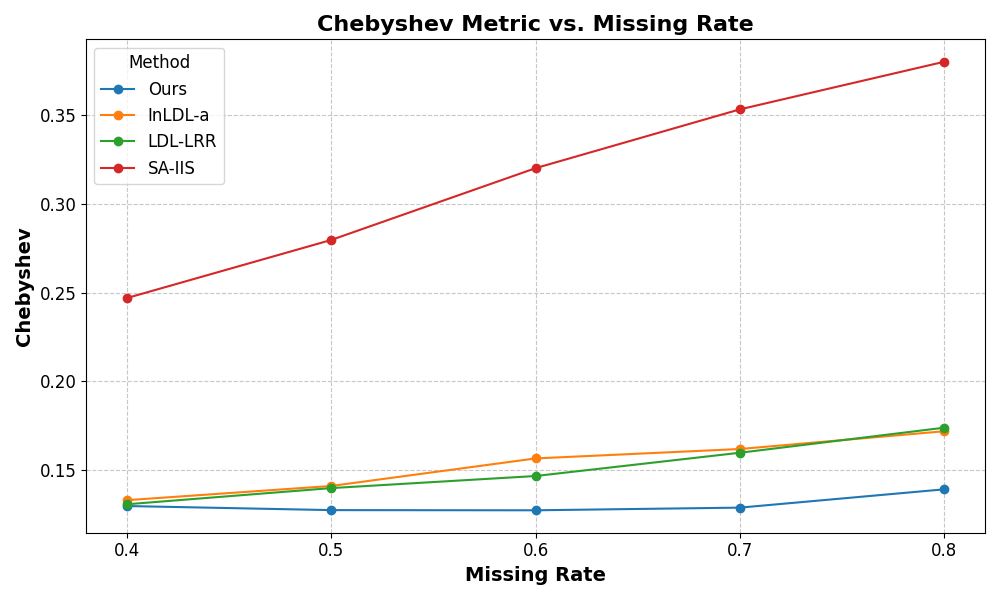}
        \caption{Movie}
        \label{Emotion6}
    \end{subfigure}
    \hspace{0.05\textwidth} 
    \begin{subfigure}[b]{0.45\textwidth} 
        \centering
        \includegraphics[width=\textwidth]{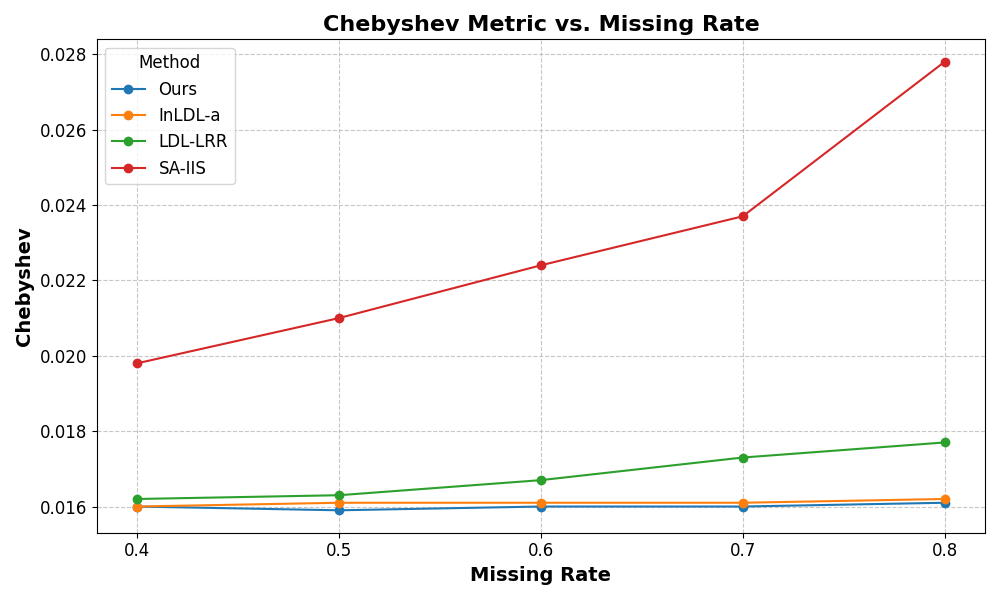}
        \caption{Yeast-elu}
        \label{SBU_3DFE}
    \end{subfigure}
    
    \vspace{0.2cm} 
    
    \begin{subfigure}[b]{0.45\textwidth} 
        \centering
        \includegraphics[width=\textwidth]{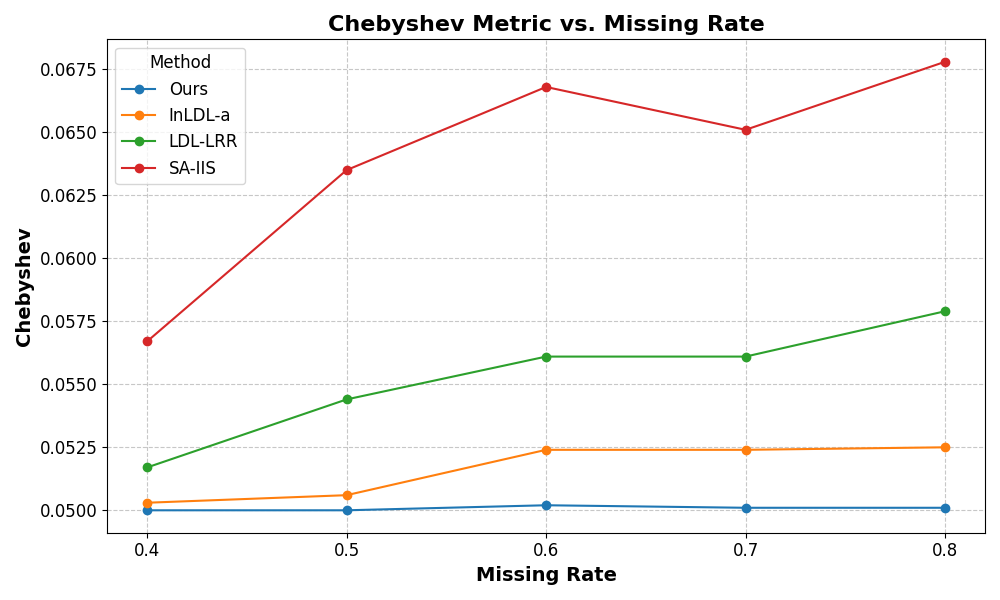}
        \caption{Yeast-cold}
        \label{Scene}
    \end{subfigure}
    \hspace{0.05\textwidth} 
    \begin{subfigure}[b]{0.45\textwidth} 
        \centering
        \includegraphics[width=\textwidth]{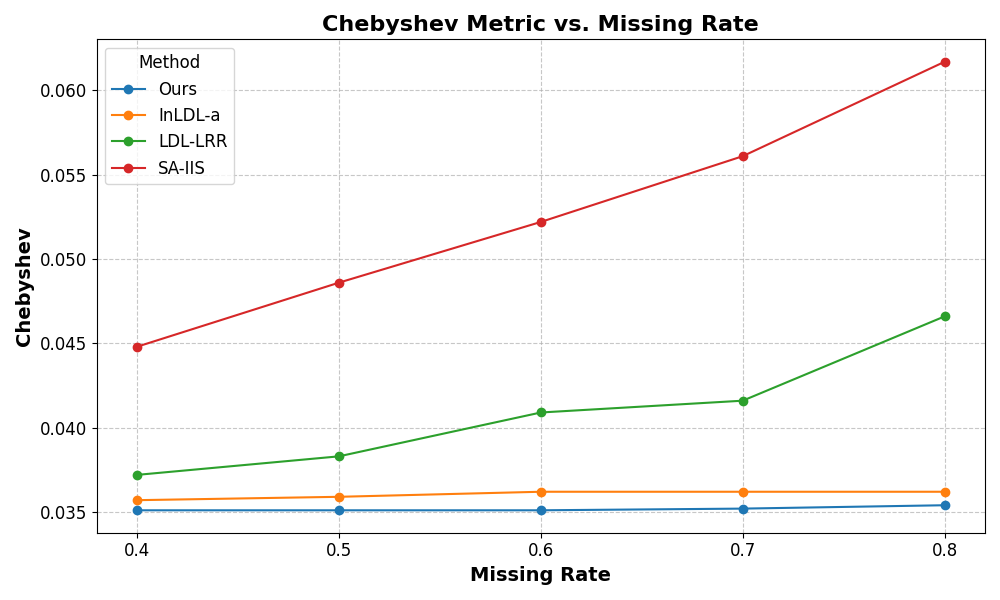}
        \caption{Yeast-dtt}
        \label{Yeast_spo}
    \end{subfigure}
    
    \caption{Predictive performance comparison under different missing rates.}
    \label{fig:images_MRandPer2}
\end{figure*}

Fig. \ref{fig:images_MRandPer2} shows the results of 4 methods in the predictive experiment under different missing rates, and the results clearly indicate that our method is less affected by missing rates. Across all four datasets, it is observed that as the missing rate increases, the performance of our method declines at a relatively slow pace. However, when it comes to InLDL-a, on Movie and Yeast-cold, the InLDL-a method exhibits a more rapid rate of performance degradation.

\section{Further Ablation Study}
 We find that in most cases, the constraint we add to maintain the proportion of observed labels is the most effective. However, as shown in Table \ref{Ablation_80}, in rare cases, the trace norm term is more effective, as evidenced by the performance of ``w/o $Cons$'' being better than 
``w/o''. All of these occur when the missing rate is very high, reaching 80\%. This is because when the missing rate is too high, there may be only one observable label in a sample, making it impossible to utilize proportional information. So the trace norm term can play a important role in extremely high missing rates.

\section{Convergence Analysis}
Here we provide a proof of convergence of Algorithm 1.

If $\{\mathbf{\Lambda}_k,\mathbf{\Lambda}'_k\}$ are bounded and 
\begin{equation}
\sum_{k=0}^{\infty}{\left( \lVert \mathbf{\Lambda} _{k+1}-\mathbf{\Lambda} _k \rVert _{\text{F}}^{2}+\lVert \mathbf{\Lambda }'_{k+1}-\mathbf{\Lambda} '_k \rVert _{\text{F}}^{2} \right) <\infty},   
\label{1}
\end{equation}
then the sequence $\{\mathbf{Z},\mathbf{\Lambda},\mathbf{\Lambda}'\}$ will converge.

Optimization Problem:
\begin{equation}
\label{equivalent_object}
\begin{aligned}
&\min _{\mathbf{A,B,D}} \quad\frac{1}{2} \operatorname{tr}\left(\mathbf{D}^{\top} \mathbf{G}\mathbf{D}\right)+\alpha\|\mathbf{A}\|_* \\
&\text { s.t. } \quad\mathbf{D}\times \mathbf{1}_{m} = \mathbf{1}_{n} , \ \mathbf{D}\geq\mathbf{0}_{n \times m}, \ \mathbf{B} \in Cons,\\
& \quad \hspace{1.8em} \mathbf{D} - \mathbf{A} = \mathbf{0},\ \mathbf{D}-\mathbf{B}=\mathbf{0}.
\end{aligned}
\end{equation}

Augmented Lagrange multiplier equation (assume $\mathbf{Z}=\{\mathbf{A},\mathbf{B},\mathbf{D}\}$):
\begin{equation}
\label{lagarange}
\begin{aligned}
\underset{\mathbf{A},\mathbf{B},\mathbf{D}}{\min}\,\,\mathcal{L}\left( \mathbf{Z},\mathbf{\Lambda} ,\mathbf{\Lambda} ' \right)
&=\frac{1}{2}\text{tr}\left( \mathbf{D}^{\top}\mathbf{GD} \right) +\alpha \lVert \mathbf{A} \rVert _*  \\*
&\quad+\left< \mathbf{\Lambda} ,\mathbf{D}-\mathbf{A} \right> +\frac{\rho}{2}\lVert \mathbf{D}-\mathbf{A} \rVert _{\text{F}}^{2} \\*
&\quad+\left< \mathbf{\Lambda} ',\mathbf{D}-\mathbf{B} \right> +\frac{\rho}{2}\lVert \mathbf{D}-\mathbf{B} \rVert _{\text{F}}^{2}\\
\text { s.t. } \mathbf{D}\times \mathbf{1_m}&=\mathbf{1_n}, \mathbf{D}\ge\mathbf{0}_{n \times m}, \mathbf{B}\in Cons.
\end{aligned}
\end{equation}

The objective function of Eq. (\ref{lagarange}) is strongly convex w.r.t. $\mathbf{A},\mathbf{B},\mathbf{D}$, because

\begin{equation}
\begin{aligned}
\mathcal{L}\left( \mathbf{Z}, \mathbf{\Lambda} , \mathbf{\Lambda} ' \right)
&=\frac{1}{2}\text{tr}\left( \mathbf{D}^{\top}\mathbf{GD} \right) +\alpha \lVert \mathbf{A} \rVert _* \\*
&\quad+\frac{\rho}{2}\lVert \mathbf{D}-\mathbf{A}+\frac{\mathbf{\Lambda}}{\rho} \rVert _{\text{F}}^{2}-\frac{1}{2\rho}\lVert \mathbf{\Lambda} \rVert _{\text{F}}^{2}  \\*
&\quad+\frac{\rho}{2}\lVert \mathbf{D}-\mathbf{B}+\frac{\mathbf{\Lambda} '}{\rho} \rVert _{\text{F}}^{2}-\frac{1}{2\rho}\lVert \mathbf{\Lambda} ' \rVert _{\text{F}}^{2} .
\end{aligned}
\end{equation}

Consequently, we have
\begin{equation}
\mathcal{L}\left( \mathbf{A}+\Delta \mathbf{A} \right) -\mathcal{L}\left( \mathbf{A} \right) \ge \partial _{\mathbf{A}}\mathcal{L}\left( \mathbf{A} \right) ^{\top}\Delta \mathbf{A}+\rho \lVert \Delta \mathbf{A} \rVert _{\text{F}}^{2}.
\label{5}
\end{equation}

For $\mathbf{A}^*$ to be the minimizer of $\mathcal{L}(\mathbf{A}^*)$, then we have
\begin{equation}
\partial _{\mathbf{A}}\mathcal{L}\left( \mathbf{A}^* \right) ^{\top}\Delta \mathbf{A}\ge 0.  
\label{6}
\end{equation}
Combining Eq. (\ref{5}) and Eq. (\ref{6}), we have
\begin{equation}
\mathcal{L}\left( \mathbf{A}_k \right) -\mathcal{L}\left( \mathbf{A}_{k+1} \right) \ge \rho \lVert \mathbf{A}_k-\mathbf{A}_{k+1} \rVert _{\text{F}}^{2}.
\label{7}
\end{equation}

Similarly,
\begin{equation}
\mathcal{L}\left( \mathbf{B}_k \right) -\mathcal{L}\left( \mathbf{B}_{k+1} \right) \ge \rho \lVert \mathbf{B}_k-\mathbf{B}_{k+1} \rVert _{\text{F}}^{2}.
\label{8}
\end{equation}
\begin{equation}
\mathcal{L}\left( \mathbf{D}_k \right) -\mathcal{L}\left( \mathbf{D}_{k+1} \right) \ge \rho \lVert \mathbf{D}_k-\mathbf{D}_{k+1} \rVert _{\text{F}}^{2}.
\label{9}
\end{equation}

Let $\nu=\min\{1,\rho\}$ and combine Eq. (\ref{7}), Eq. (\ref{8}) and Eq. (\ref{9}), we obtain
\begin{equation}
\small
\begin{aligned}
&\mathcal{L}\left( \mathbf{Z}_k,\mathbf{\Lambda} _k,\mathbf{\Lambda} '_k \right) -\mathcal{L}\left( \mathbf{Z}_{k+1},\mathbf{\Lambda} _{k+1},\mathbf{\Lambda} '_{k+1} \right)\\
&=\mathcal{L}\left( \mathbf{Z}_k,\mathbf{\Lambda} _k,\mathbf{\Lambda} '_k \right) -\mathcal{L}\left( \mathbf{Z}_{k+1},\mathbf{\Lambda} _k,\mathbf{\Lambda} '_k \right)  \\*
&\quad+\mathcal{L}\left( \mathbf{Z}_{k+1},\mathbf{\Lambda} _k,\mathbf{\Lambda} '_k \right) -\mathcal{L}\left( \mathbf{Z}_{k+1},\mathbf{\Lambda} _{k+1},\mathbf{\Lambda} '_{k+1} \right) \\
&\ge \nu \lVert \mathbf{Z}_k-\mathbf{Z}_{k+1} \rVert _{\text{F}}^{2}-\frac{1}{\rho}\left( \lVert \mathbf{\Lambda} _k-\mathbf{\Lambda} _{k+1} \rVert _{\text{F}}^{2}+\lVert \mathbf{\Lambda} '_k-\mathbf{\Lambda} '_{k+1} \rVert _{\text{F}}^{2} \right) \\
&\ge \nu \lVert \mathbf{Z}_k-\mathbf{Z}_{k+1} \rVert _{\text{F}}^{2}-\frac{1}{\nu}\left( \lVert \mathbf{\Lambda} _k-\mathbf{\Lambda} _{k+1} \rVert _{\text{F}}^{2}+\frac{1}{\nu}\lVert \mathbf{\Lambda} '_k-\mathbf{\Lambda} '_{k+1} \rVert _{\text{F}}^{2} \right).
\end{aligned}
\end{equation}

Recalling that $\mathcal{L}\left( \mathbf{Z},\mathbf{\Lambda},\mathbf{\Lambda} ' \right) $ is bounded below, we have

\begin{equation}
\tiny
\sum_{k=0}^{\infty}{\nu \lVert \mathbf{Z}_k-\mathbf{Z}_{k+1} \rVert _{\text{F}}^{2}}-\sum_{k=0}^{\infty}{\frac{1}{\nu}\left( \lVert \mathbf{\Lambda} _k-\mathbf{\Lambda} _{k+1} \rVert _{\text{F}}^{2}+\frac{1}{\nu}\lVert \mathbf{\Lambda} '_k-\mathbf{\Lambda} '_{k+1} \rVert _{\text{F}}^{2} \right)}<\infty.
\label{11}
\end{equation}

We have assumed the second term is bounded (in Eq. (\ref{1})), so we can immediately get
\begin{equation}
\sum_{k=0}^{\infty}{\nu \lVert \mathbf{Z}_k-\mathbf{Z}_{k+1} \rVert _{\text{F}}^{2}}<\infty.
\end{equation}

Recalling $\mathbf{Z}=\{\mathbf{A},\mathbf{B},\mathbf{D}\}$,we get the \textbf{conclusion}: As the training progresses, ${\mathbf{A},\mathbf{B},\mathbf{D}\ } $ in Eq. (9) will converge. i.e., $\mathbf{A}_{k+1}-\mathbf{A}_k \rightarrow 0$.

\section{Proof of Theorem 1}

\textbf{Theorem 1.} \textit{Under certain assumptions, the error of the scaling coefficient can be expressed as}

\begin{equation}
\small
\label{Theorem1}
    k^g_i-k_i = \frac{1-\Sigma_k d_{ik}M_{ik}}{\Sigma_k d^g_{ik}M_{ik}}.
\end{equation}

\textbf{Proof:}

From Eq. (3) and Eq. (15) in the paper, we have

\begin{equation}
\label{kgi}
\begin{aligned}
    k^g_i&=\frac{\mathit{D^{\mathit{o}}_{\mathit{ij}}}}{\mathit{D^{\mathit{g}}_{\mathit{ij}}}}=\frac{1}{\sum_{k=1}^m \mathit{D^{\mathit{g}}_{\mathit{ik}}} \cdot \mathit{M_{\mathit{ik}}}}
\end{aligned}
\end{equation}

and

\begin{equation}
\label{ki}
    k_i=\dfrac{\sum_{k=1}^m \left(\rho D_{i k} + \Lambda'_{ i k}\right) \cdot D^o_{ik} \cdot M_{i k}}{\rho \sum_{k=1}^m(D^o_{ik})^2 \cdot M_{i k}}.
\end{equation}

Consider the error of the scaling coefficients:

\begin{equation}
\label{first}
    \begin{aligned}
        k^g_i-k_i&=\frac{1}{\sum_{k=1}^m\mathit{D^{\mathit{g}}_{\mathit{ik}}} \mathit{M_{\mathit{ik}}}}-\dfrac{\sum_{k=1}^m \left( D_{i k} + \frac{\Lambda'_{ i k}}{\rho}\right) D^o_{ik}  M_{i k}}{ \sum_{k=1}^m(D^o_{ik})^2 M_{i k}}\\
        &\leq\frac{1}{\sum_{k=1}^m \mathit{D^{\mathit{g}}_{\mathit{ik}}}\mathit{M_{\mathit{ik}}}}-\dfrac{\sum_{k=1}^m \left( D_{i k} + \frac{\Lambda'_{ i k}}{\rho}\right)  M_{i k}}{ \sum_{k=1}^mD^o_{ik}  M_{i k}}
    \end{aligned}
\end{equation}

Assume that $\sigma_i$, i.e. the sum of the labels in $i$-th instance that are masked is relatively small, there holds $\mathit{D^{\mathit{g}}_{\mathit{ik}}}\approx D^o_{ik}$, and the above equation can be further rewritten as

\begin{equation}
\label{first}
    \begin{aligned}
        k^g_i-k_i&\leq\frac{1}{\sum_{k=1}^m\mathit{D^{\mathit{g}}_{\mathit{ik}}} \mathit{M_{\mathit{ik}}}}-\dfrac{\sum_{k=1}^m \left( D_{i k} + \frac{\Lambda'_{ i k}}{\rho}\right)  M_{i k}}{ \sum_{k=1}^mD^g_{ik} M_{i k}}\\
        &=\frac{1-\sum_{k=1}^m D_{ik} M_{ik}-\sum_{k=1}^m \frac{\Lambda'_{ i k}}{\rho}}{\sum_{k=1}^m D^g_{ik}\ M_{ik}}
    \end{aligned}
\end{equation}

In the above equation, $\frac{\Lambda'_{ i k}}{\rho}$ can be considered as the gradient of $D_{ik}$, and should be equal to zero after optimization. So the final equation becomes

\begin{equation}
\label{first}
    \begin{aligned}
        k^g_i-k_i\leq\frac{1-\sum_{k=1}^m D_{ik} M_{ik}}{\sum_{k=1}^m D^g_{ik} M_{ik}},
    \end{aligned}
\end{equation}

so \textbf{Theorem 1} in the paper is proved.



\begin{table*}[ht]
\centering
\setlength{\tabcolsep}{1mm}
\begin{tabular}{cccccccc}
\hline
Datasets & Ours &  InLDL-a  & WInLDL & SA-IIS & LDL-LRR & PT-Bayes & LDL-DPA\\
\hline
alpha &
\textbf{.0106 ± .0001}&
\underline{.0136 ± .0000}&
.0171 ± .0003&
.0169 ± .0001&
.0138 ± .0001&
.0356 ± .0015&
.0177 ± .0003\\ 
cdc &
\textbf{.0117 ± .0001}&
\underline{.0164 ± .0001}&
.0208 ± .0002&
.0204 ± .0003&
.0165 ± .0002&
.0394 ± .0017&
.0214 ± .0002\\
cold &
\textbf{.0386 ± .0001}&
\underline{.0520 ± .0003}&
.0596 ± .0010&
.0625 ± .0011&
.0536 ± .0010&
.0993 ± .0110&
.0659 ± .0019\\
dtt &
\textbf{.0273 ± .0001}&
\underline{.0370 ± .0003}&
.0469 ± .0013&
.0498 ± .0015&
.0390 ± .0012&
.0932 ± .0082&
.0538 ± .0008\\
elu &
\textbf{.0120 ± .0001}&
\underline{.0163 ± .0000}&
.0213 ± .0005&
.0208 ± .0005&
.0166 ± .0001&
.0411 ± .0029&
.0217 ± .0006\\
spo &
\textbf{.0403 ± .0009}&
\underline{.0584 ± .0001}&
.0634 ± .0007&
.0638 ± .0012&
.0589 ± .0003&
.0913 ± .0027&
.0667 ± .0008\\
SJAFFE & 
\textbf{.0785 ± .0022}&
.1143 ± .0078&
.1882 ± .0108&
\underline{.1089 ± .0048}&
.1203 ± .0011&
.1256 ± .0057&
.2657 ± .0149\\
Scene &
\textbf{.2760 ± .0035}&
\underline{.3276 ± .0025}&
.3371 ± .0029&
.3299 ± .0025&
.3483 ± .0011&
.6317 ± .0128&
.3330 ± .0032\\
Movie 
& \textbf{.0916 ± .0003}&
\underline{.1301 ± .0008}&
.1822 ± .0013&
.2298 ± .0017&
.1357 ± .0014&
.7735 ± .0249&
.2090 ± .0072\\
SBU & \textbf{.0982 ± .0011}&
.1272 ± .0012&
\underline{.1134 ± .0009}&
.1322 ± .0004&
.1368 ± .0004&
.1331 ± .0009&
.1291 ± .0010\\
Emo & \textbf{.2376 ± .0025}&
\underline{.3061 ± .0011}&
.3121 ± .0024&
.3201 ± .0022&
.3204 ± .0019&
.3220 ± .0020&
.3166 ± .0020\\
RAF &
.2715 ± .0022&
.3923 ± .0016&
.6962 ± .0009&
\textbf{.2457 ± .0032}&
.2707 ± .0012&
.2963 ± .0155&
\underline{.2462 ± .0027}\\
\hline
\end{tabular}
\caption{Chebyshev (the lower the better) results for the \textbf{recovery} setting on all datasets when missing rate $\omega$ = 50\%.  InLDL-a is the abbreviation for IncomLDL-admm. The value is shown in mean±std form. Bold and underlined indicate the best and second best results, respectively.}
\label{Chebyshev}
\end{table*}

\begin{table*}[ht]
    \centering
    \setlength{\tabcolsep}{1mm}
    \begin{tabular}{cccccccc} \hline
    Dataset & Ours & InLDL-a & WInLDL & SA-IIS & LDL-LRR & PT-Bayes & LDL-DPA  \\ \hline
       alpha & \textbf{0.1945 ± .0010} & \underline{0.2146 ± .0004} & 0.9080 ± .0169 & 0.4548 ± .0091 & 0.2382 ± .0090 & 0.9399 ± .0663 & 0.5016 ± .0240 \\ 
        cdc & \textbf{0.1966 ± .0019} & \underline{0.2187 ± .0004} & 0.7998 ± .0576 & 0.4209 ± .0171 & 0.2398 ± .0063 & 0.8023 ± .0565 & 0.4809 ± .0120 \\ 
        cold & \textbf{0.1349 ± .0006} & \underline{0.1453 ± .0019} & 0.3034 ± .0113 & 0.1977 ± .0083 & 0.1515 ± .0027 & 0.3379 ± .0322 & 0.2059 ± .0094 \\ 
        dtt & \textbf{0.0958 ± .0004} & \underline{0.1009 ± .0007} & 0.2632 ± .0138 & 0.1551 ± .0091 & 0.1129 ± .0055 & 0.3060 ± .0211 & 0.1727 ± .0090 \\ 
        elu & \textbf{0.1830 ± .0008} & \underline{0.2039 ± .0006} & 0.7945 ± .0274 & 0.4082 ± .0077 & 0.2272 ± .0062 & 0.7924 ± .0535 & 0.4509 ± .0114 \\ 
        spo & \textbf{0.2315 ± .0009} & \underline{0.2548 ± .0015} & 0.4741 ± .0157 & 0.3181 ± .0046 & 0.2623 ± .0049 & 0.4991 ± .0385 & 0.3301 ± .0083 \\ 
        SJAFFE & 0.4091 ± .0144 & 0.4296 ± .0026 & \underline{1.5932 ± .0125} & 0.4966 ± .0196 & 0.4346 ± .0158 & 0.5585 ± .0355 & \textbf{1.5829 ± .0480} \\ 
        Scene & \textbf{2.4534 ± .0009} & 2.4718 ± .0009 & 2.5182 ± .0039 & \underline{2.4707 ± .0013} & 2.4707 ± .0010 & 2.4965 ± .0341 & 2.4968 ± .0048 \\ 
        Movie & \textbf{0.6130 ± .0029} & 0.7146 ± .0077 & 1.0812 ± .0070 & 1.2111 ± .0038 & \underline{0.7123 ± .0081} & 1.5170 ± .0846 & 0.9781 ± .0130 \\ 
        SBU & \textbf{0.3954 ± .0023} & 0.4129 ± .0018 & 0.6107 ± .0103 & 0.4134 ± .0034 & \underline{0.4125 ± .0006} & 0.4198 ± .0040 & 0.4393 ± .0037 \\ 
        Emo & \textbf{1.6046 ± .0024} & \underline{1.6895 ± .0042} & 1.7651 ± .0092 & 1.7191 ± .0069 & 1.7025 ± .0058 & 1.7355 ± .0297 & 1.7238 ± .0046 \\ 
        RAF & 1.6233 ± .0021 & 2.1085 ± .0050 & 2.2144 ± .0081 & \underline{1.5751 ± .0076} & \textbf{1.5723 ± .0015} & 1.5881 ± .0183 & 1.5765 ± .0028 \\ \hline
    \end{tabular}
     \caption{Clark (the lower the better) results for the \textbf{recovery} setting on all datasets when missing rate $\omega$ = 80\%. InLDL-a is the abbreviation for IncomLDL-admm. The value is shown in mean±std form. Bold and underlined indicate the best and second best results, respectively.}
    \label{clark-80-recovery}
\end{table*}

\begin{table*}[ht]
    \centering
     \setlength{\tabcolsep}{1mm}
     \begin{tabular}{cccccccc} \hline 
        Dataset & Ours & InLDL-a & WInLDL & SA-IIS & LDL-LRR & PT-Bayes & LDL-DPA \\ \hline
        alpha & \textbf{.0135 ± .0001} & \underline{.0136 ± .0000} & .0178 ± .0002 & .0169 ± .0002 & .0136 ± .0001 & .0339 ± .0029 & .0187 ± .0003 \\ 
        cdc & \textbf{.0164 ± .0001} & \underline{.0164 ± .0001} & .0223 ± .0004 & .0208 ± .0003 & .0168 ± .0002 & .0440 ± .0035 & .0231 ± .0003 \\ 
        cold & \textbf{.0501 ± .0001} & \underline{.0510 ± .0005} & .0629 ± .0026 & .0612 ± .0017 & .0533 ± .0009 & .1114 ± .0194 & .0679 ± .0028 \\ 
        dtt & \textbf{.0350 ± .0001} & \underline{.0360 ± .0003} & .0507 ± .0030 & .0507 ± .0033 & .0414 ± .0028 & .0983 ± .0105 & .0583 ± .0048 \\ 
        elu & \textbf{.0159 ± .0000} & \underline{.0160 ± .0000} & .0226 ± .0006 & .0210 ± .0006 & .0164 ± .0002 & .0444 ± .0048 & .0234 ± .0007 \\ 
        spo & \textbf{.0592 ± .0001} & \underline{.0593 ± .0003} & .0674 ± .0015 & .0649 ± .0017 & .0599 ± .0005 & .0934 ± .0041 & .0708 ± .0016 \\ 
        SJAFFE & \textbf{.1107 ± .0040} & .1197 ± .0049 & .2595 ± .0043 & \underline{.1192 ± .0054} & .1211 ± .0030 & .1344 ± .0045 & .2303 ± .0206 \\ 
        Scene & \textbf{.3371 ± .0036} & .3492 ± .0034 & .3598 ± .0047 & .3496 ± .0028 & .3548 ± .0017 & .6087 ± .0196 & \underline{.3490 ± .0039} \\ 
        Movie & \textbf{.1270 ± .0004} & .1419 ± .0005 & .2558 ± .0029 & .2815 ± .0037 & \underline{.1390 ± .0014} & .7793 ± .0134 & .2186 ± .0042 \\ 
        SBU3DFE & \underline{.1275 ± .0012} & .1330 ± .0011 & \textbf{.1238 ± .0015} & .1374 ± .0011 & .1398 ± .0005 & .1367 ± .0007 & .1340 ± .0016 \\ 
        Emo & \textbf{.3014 ± .0012} & .3121 ± .0017 & .3169 ± .0029 & .3727 ± .0045 & .3223 ± .0035 & \underline{.3107 ± .0032} & .3283 ± .0054 \\ 
        RAF & .2819 ± .0018 & \textbf{.2196 ± .0014} & .5538 ± .0055 & .2670 ± .0035 & .2714 ± .0021 & .2797 ± .0059 & \underline{.2546 ± .0019} \\ \hline
    \end{tabular}
    \caption{Chebyshev (the lower the better) results for the \textbf{predictive} setting on all datasets when missing rate $\omega$ = 50\%. InLDL-a is the abbreviation for IncomLDL-admm. The value is shown in mean±std form. Bold and underlined indicate the best and second best results, respectively.}
    \label{cheby-50-predictive}
\end{table*}

\begin{table*}[ht]
    \centering
    \setlength{\tabcolsep}{1mm}
    \begin{tabular}{cccccccc} \hline 
        Dataset & Ours & InLDL-a & WInLDL & SA-IIS & LDL-LRR & PT-Bayes & LDL-DPA  \\ \hline
        alpha & \textbf{0.2136 ±  .0015} & \underline{0.2159 ±  .0009} & 0.8084 ±  .0266 & 0.3836 ±  .0090 & 0.2390 ±  .0090 & 0.8664 ±  .0614 & 0.5579 ±  .0198  \\ 
        cdc & \textbf{0.2158 ±  .0005} & \underline{0.2181 ±  .0006} & 0.7939 ±  .0373 & 0.3900 ±  .0111 & 0.2486 ±  .0076 & 0.8847 ±  .0291 & 0.5273 ±  .0160  \\ 
        cold & \textbf{0.1354 ±  .0005} & \underline{0.1403 ±  .0012} & 0.3050 ±  .0157 & 0.1769 ±  .0096 & 0.1492 ±  .0059 & 0.3430 ±  .0498 & 0.2244 ±  .0146  \\ 
        dtt & \textbf{0.0962 ±  .0003} & \underline{0.0976 ±  .0003} & 0.2897 ±  .0121 & 0.1520 ±  .0099 & 0.1176 ±  .0048 & 0.3037 ±  .0543 & 0.1872 ±  .0089  \\ 
        elu & \textbf{0.1954 ±  .0004} & \underline{0.1989 ±  .0011} & 0.7379 ±  .0249 & 0.3607 ±  .0101 & 0.2271 ±  .0113 & 0.8220 ±  .0632 & 0.5033 ±  .0204  \\ 
        spo & \textbf{0.1276 ±  .0702} & \underline{0.1293 ±  .0711} & 0.4749 ±  .0269 & 0.3046 ±  .0082 & 0.2684 ±  .0048 & 0.5092 ±  .0755 & 0.3572 ±  .0157  \\ 
        SJAFFE & \textbf{0.4306 ±  .0108} & \underline{0.4387 ±  .0043} & 1.3074 ±  .0488 & 0.4767 ±  .0280 & 0.4481 ±  .0166 & 0.5976 ±  .0444 & 1.4876 ±  .1105  \\ 
       Scene & \textbf{2.4735 ±  .0017} & \underline{2.4807 ±  .0016} & 2.5281 ±  .0034 & 2.4840 ±  .0019 & 2.4809 ±  .0014 & 2.5186 ±  .0187 & 2.5204 ±  .0054  \\ 
        Movie & \textbf{0.6159 ±  .0043} & \underline{0.6875 ±  .0216} & 1.0864 ±  .0141 & 1.2592 ±  .0254 & 0.7175 ±  .0070 & 1.4884 ±  .1344 & 0.9911 ±  .0249  \\ 
        SBU& \textbf{0.4085 ±  .0007} & 0.4214 ±  .0055 & 0.5126 ±  .0302 & 0.4189 ±  .0042 & \underline{0.4183 ±  .0031} & 0.4260 ±  .0065 & 0.4483 ±  .0099  \\ 
        Emo & \textbf{1.6874 ±  .0155} & 1.7139 ±  .0366 & 1.7416 ±  .0195 & 1.8348 ±  .0213 & 1.7227 ±  .0077 & \underline{1.6961 ±  .0157} & 1.7497 ±  .0078  \\ 
        RAF & 1.5887 ±  .0014 & 1.5872 ±  .0143 & 1.7471 ±  .0078 & 1.5817 ±  .0089 & \textbf{1.5737 ±  .0046} & \underline{1.5737 ±  .0024} & 1.5856 ±  .0064  \\ \hline
    \end{tabular}
    \caption{Clark (the lower the better) results for the \textbf{predictive} setting on all datasets when missing rate $\omega$ = 80\%. InLDL-a is the abbreviation for IncomLDL-admm. The value is shown in mean±std form. Bold and und
     erlined indicate the best and second best results, respectively.}
     \label{clark-80-predictive}
\end{table*}

\begin{table*}[htbp]
    \centering \caption{Ablation Results when missing rate $\omega=80 \%$ on 4 datasets. $\uparrow$ ($\downarrow$) indicates the higher (lower) the better.}
    \resizebox{0.96\textwidth}{!}{%
    \begin{tabular}{c|c|cccc|cccc}\hline 
         \multirow{2}{*}{\centering Metric} & \multirow{2}{*}{\centering Method} & \multicolumn{4}{c|}{Recovery} & \multicolumn{4}{c}{Predictive} \\ 
    \cline{3-10}
    & &Yeast-dtt & Yeast-elu & Yeast-spo & Emotion6 & Yeast-dtt & Yeast-elu & Yeast-spo & Emotion6 \\
        \hline
        ~ & Ours & \textbf{0.0958} & \textbf{0.1834} & \textbf{0.2299} & \textbf{1.6009} &\textbf{0.0960} & \textbf{0.1957} & \textbf{0.2532} & \textbf{1.6635}\ \\ 
        $clark\downarrow$ & w/o $Cons$ &0.1014 & 0.2043 & 0.2559 & 1.6789   &0.0979 & 0.1994 & 0.2565 & 1.6664\\ 
        ~ & w/o TN &0.1871 & 0.1943 & 0.2653 & 1.8921  &0.1169 & 0.1999 &0.2638 & 1.7109\\
        \hline
        ~ & Ours & \textbf{0.9942} & \textbf{0.9949} & \textbf{0.9797} & \textbf{0.7081} &\textbf{0.9944}&\textbf{0.9942}&\textbf{0.9763}&\textbf{0.7071}\\ 
        $cosine\uparrow$& w/o $Cons$ &0.9937  &0.9938  &0.9747  &0.6566  &0.9941&0.9941&0.9749&0.6660 \\ 
        ~ & w/o TN &0.9782  &0.9943  & 0.9732 &0.4824  &0.9918&0.9940&0.9738&0.6290\\ 
        \hline
    \end{tabular}
    }
    \label{Ablation_80}
\end{table*}

\end{document}